\documentclass[lettersize,journal]{IEEEtran}
\usepackage[utf8]{inputenc}
\usepackage{amsmath,amsfonts}
\usepackage{algorithmic}
\usepackage{algorithm}
\usepackage{array}
\usepackage[caption=false,font=normalsize,labelfont=sf,textfont=sf]{subfig}
\usepackage{textcomp}
\usepackage{stfloats}
\usepackage{url}
\usepackage{verbatim}
\usepackage{graphicx}
\usepackage{cite}
\usepackage{booktabs}
\usepackage{multirow}%
\usepackage{tabularray}
\DeclareUnicodeCharacter{FF0C}{\textvisiblespace}

\begin{document}
\title{Systematic Abductive Reasoning via Diverse Relation Representations in Vector-symbolic Architecture}{

\author{\IEEEauthorblockN{Zhong-Hua Sun, Ru-Yuan Zhang, Zonglei Zhen, Da-Hui Wang, Yong-Jie Li, Xiaohong Wan, Hongzhi You*}

\thanks{Corresponding author: Hongzhi~You, email: hongzhi-you@uestc.edu.cn}%
\thanks{Zhong-Hua~Sun, Yong-Jie Li and Hongzhi~You are with School of Life Science and Technology, University of Electronic Science and Technology of China (UESTC), Chengdu, China. Ru-Yuan Zhang is with Brain Health Institute, National Center for Mental Disorders, Shanghai Mental Health Center, Shanghai Jiao Tong University School of Medicine and School of Psychology, Shanghai, China. Zonglei Zhen and Xiaohong Wan are with State Key Laboratory of Cognitive Neuroscience and Learning, Beijing Normal University, Beijing， China. Da-Hui Wang is with School of Systems Science, Beijing Normal University, Beijing, China. }}

}



\maketitle

\begin{abstract}


In abstract visual reasoning, monolithic deep learning models suffer from limited interpretability and generalization, while existing neuro-symbolic approaches fall short in capturing the diversity and systematicity of attributes and relation representations. To address these challenges, we propose a Systematic Abductive Reasoning model with diverse relation representations (Rel-SAR) in Vector-symbolic Architecture (VSA) to solve Raven's Progressive Matrices (RPM). To derive attribute representations with symbolic reasoning potential, we introduce not only various types of atomic vectors that represent numeric, periodic and logical semantics, but also the structured high-dimentional representation (SHDR) for the overall Grid component. For systematic reasoning, we propose novel numerical and logical relation functions and perform rule abduction and execution in a unified framework that integrates these relation representations. Experimental results demonstrate that Rel-SAR achieves significant improvement on RPM tasks and exhibits robust out-of-distribution generalization. Rel-SAR leverages the synergy between HD attribute representations and symbolic reasoning to achieve systematic abductive reasoning with both interpretable and computable semantics.
 
\end{abstract}

\section{Introduction}


Raven's Progressive Matrices (RPM) are a family of psychological intelligence tests widely used for the assessment of abstract reasoning \cite{carpenter1990one, bilker2012development}. From a cognitive psychology perspective, abstract visual reasoning in RPM tests involves constructing high-level representations from images and deriving potential relations from these representations \cite{carpenter1990one, mitchell2021abstraction}. Endowing artificial intelligence with such capabilities is now regarded as a crucial step toward achieving human-level intelligence. However, many recent monolithic deep learning models, which do not explicitly separate perception and reasoning \cite{barrett2018measuring, hill2019learning, zhang2019learning, zheng2019abstract, hu2021stratified, benny2021scale}, face inherent challenges, such as poor interpretability, limited robustness and generalization, and difficulties in module reuse \cite{zhang2021abstract}. Neuro-symbolic architecture, which combines neural visual perception with symbolic reasoning, offers a promising approach to overcoming these challenges and achieving human-level interpretability and generalization \cite{marcus2003algebraic, zhang2021abstract, hersche2023neuro}. 

In neuro-symbolic architectures (NSA), Marcus argues that symbol-manipulation in cognition involves representing relations between variables \cite{marcus2003algebraic}. For RPM tests, object attributes serve as the variables, while potential rules involve the relations. Nevertheless, due to incomplete attribute and relation representations, achieving systematic abduction and execution is still a critical challenge for NSA when performing RPM tests. From the perspective of attributes, recent models such as PrAE \cite{zhang2021abstract}, the ALANS learner \cite{zhang2022learningb}, and NVSA (neuro-vector-symbolic architecture) \cite{hersche2023neuro} construct attribute representations through neural perception frontends. Notably, the NVSA model achieves hierarchically structured VSA representations of image panels, capturing multiple objects with multiple attributes \cite{hersche2023neuro}. Regarding relation representations, PrAE and NVSA achieve abstract reasoning through probabilistic abduction and execution \cite{zhang2021abstract} and distributed vector-symbolic architecture (VSA) \cite{hersche2023neuro}, respectively. Both models rely on predetermined multiple rule templates, each specialized for distinct individual RPM rules. To address the limitations in rule expressiveness, the ALANS learner utilizes learnable rule operators in the abstract algebraic structure, without manual definition for every rules \cite{zhang2022learningb}. Additionally, the ARLC model adopts a more expressive VSA-based rule template, operating in the rule parameter space \cite{camposampiero2024towards}. Both models offer improved interpretabiltiy and generalizability. Despite their advances, previous models fall short in capturing the diversity and systematicity of attribute and relation representations. In contrast, human cognition demonstrates rich and flexible internal representations \cite{mansouri2020emergence, marcus2020insights}, including arithmetic and logic, and rule-based reasoning systems in cognition are productive and systematic \cite{sloman1996empirical}. Therefore, the abstract visual reasoning performance of these models remains open to further improvement.

Previous research indicates that Vector Symbolic Architecture (VSA), a form of high-dimensional (HD) distributed representation, possesses algebraic properties for mathematical operations and can also achieve structured symbolic representations of data \cite{plate1994distributed,Frady_Kleyko_Kymn_Olshausen_Sommer_2021, kleyko2022survey}. In this work, to achieve comprehensive relation representations, we introduce various types of VSA-based atomic HD vectors with distinct semantic representations, including numeric values, periodic values, and logical values. Given that reasoning in RPM problems involves the overall attributes of multiple objects, we further introduce the structured HD representation (SHDR) for the \textsf{nxn Grid}. They serve as attribute representations necessary for abductive reasoning. Meanwhile, we propose numerical and logical relation functions as relation representations that take multiple HD attribute representations as input and define relations among them. Unlike rule templates designed for individual rules, the two proposed relation functions are specifically tailored to numerical and logical types, providing strong rule expressiveness.  

Here, we propose a Systematic Abductive Reasoning model with diverse relation representations (Rel-SAR) for solving RPM, inspired by the original NVSA model \cite{hersche2023neuro}. In the Rel-SAR model, visual attribute extraction and rule inference are implemented within a fully unified computational framework in the VSA machinery. The model comprises a neuro-vector frontend for perceiving object attributes of all raw images in RPM problems and a generic vector-symbolic backend for achieving symbolic reasoning. The perception frontend operates on scene-based SHDR of each image panel, which contains multiple objects, each with various attributes, and predicts HD attribute representations by VSA-based symbolic manipulations. The reasoning backend implements the core idea of systematic abductive reasoning: if the given attributes in an RPM adhere to a specific numerical or logical rule, then the relation representations of all attribute pairs can be defined using the corresponding relation functions with identical parameters. These diverse relation representations are involved in both rule abduction and execution phases, enhancing interpretability and improving the capacity for systematic abductive reasoning.

\section{Related Work}

\begin{figure}[ht]
	\centering
	\includegraphics[width=1\linewidth]{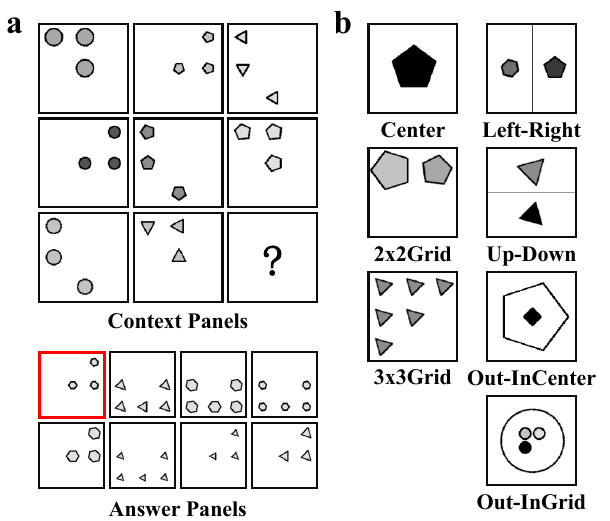} 
	\caption{\textbf{Illustrations for RAVEN dataset.} 
		\textbf{(a)} An example of RPM test from RAVEN\cite{zhang2019raven} dataset. In an RPM test, there are 8 context panels and 8 candidate panels. Participants are required to identify the underlying rules governing various attributes within the context panels. Subsequently, participants use these rules to infer the attributes of the missing panel (represented by "?") and choose the most appropriate option (highlighted with a red box) from the answer panels. \textbf{(b)} The RAVEN dataset includes seven configurations: Center, 2x2Grid, 3x3Grid, Left-Right (L-R), Up-Down (U-D), Out-InCenter (O-IC) and Out-InGrid (O-IG) \cite{zhang2019raven}. Four types of rules, i.e., \textit{Constant}, \textit{Progression}, \textit{Arithmetic}, and \textit{Distribute Three}, are applied to five attributes, i.e., \textit{Position}, \textit{Number}, \textit{Type}, \textit{Size}, and \textit{Color}, in a row-wise manner. The I-RAVEN dataset \cite{hu2021stratified} is a variant of RAVEN, where answer sets are generated using an attribute bisection tree.
	}
	\label{fig:RAVENexample}
\end{figure}


The Raven Progressive Matrices (RPM) is a widely used nonverbal intelligence test designed to assess abstract reasoning. To explore the limitations of current machine learning approaches in solving abstract reasoning tasks, two automatically generated RPM-based datasets—RAVEN \cite{zhang2019raven} and I-RAVEN \cite{hu2021stratified}—have been introduced (Figure \ref{fig:RAVENexample}). Early efforts on RPM primarily employed Relation Network (RN) \cite{santoro2017simple} and their variants \cite{barrett2018measuring, jahrens2020solving, zheng2019abstract, benny2021scale} to extract relations between context panels. Concurrently, CoPINet \cite{zhang2019learning}, MLCL \cite{malkinski2022multi}, and DCNet \cite{zhuo2022effective} integrate contrastive learning in their models. Approaches like MRNet \cite{benny2021scale} and DRNet \cite{zhao2024learning} aimed to enhance perception capabilities, while SRAN \cite{hu2021stratified} and PredRNet \cite{yang2023neural} abstract relations using stratified models and prediction errors, respectively. In addition, several methods have focused on scene decomposition and feature disentanglement \cite{wu2020scattering, mondal2023learning, spratley2020closer}. Although these monolithic deep learning models achieve high accuracy, they often suffer from limited interpretability and systematic generalization capabilities.

Another branch for solving RPM is based on neuro-symbolic architectures, which explicitly distinguish between perception and reasoning. PrAE \cite{zhang2021abstract} employs an object CNN to generate probabilistic scene representations and uses predetermined rule templates for probabilistic abduction and execution. Inspired by abstract algebra and representation theory, ALANS \cite{zhang2022learningb}, which shares the same perception frontend as PrAE, transforms probabilistic scene distributions into matrix-based algebraic representations. The algebraic reasoning backend of ALANS induces potential rules through trainable operator matrices, eliminating the need for manual rule definitions. In abstract reasoning, Vector Symbolic Architectures (VSA) serve as a bridge between perception and reasoning modules by leveraging its structured distribution representations and algebraic properties. NVSA \cite{hersche2023neuro} projects each RPM panel into a high-dimensional vector using a trainable CNN and derives probability mass functions (PMFs) by querying an external codebook. Its reasoning backend embeds these PMFs into distributed VSA representations and performs rule abduction and execution using templates based on VSA algebraic operations. NVSA  provides a differentiable and transparent implementation of probabilistic abductive reasoning by leveraging VSA representations and operators. However, its perception frontend requires searching a large external codebook, and its reasoning backend still relies on predetermined rule templates. In contrast, Learn-VRF \cite{hersche2024probabilistic}, focuses on reasoning by learning VSA rule formulations, eliminating the need for predetermined templates. ARLC \cite{camposampiero2024towards} further enhances reasoning by incorporating context augmentation and extending rule templates to accommodate more diverse rules. While ARLC and Learn-VRF implement systematic rule learning, they still struggle to process all RPM rules due to limitations in attribute representation. Recently, a class of methods known as relational bottlenecks has been proposed to enable efficient abstraction, but their capacity to handle complex relations remains uncertain\cite{webb2020emergent,altabaa2023abstractors, kerg2022neural,webb2024relational}. To address this limitation, Rel-SAR transforms perceptual inputs into high-dimensional attribute representations with symbolic reasoning potential and abducts both logical and numerical rules within a unified framework.

\section{Preliminaries}
\subsection{VSA models utilized in this study}
\noindent VSAs are a class of computational models that utilize high-dimensional distributed representations \cite{kleyko2022survey}. VSA models used in this study are Holographic Reduced Representations (HRR) and its form in the frequency domain, referred to as Fourier Holographic Reduced Representations (FHRR)\cite{plate1995holographic}. A random FHRR atomic vector, denoted as $\boldsymbol{\theta}:=\left\{ \theta _i \right\} _{i=1}^{d}$, is composed of elements $\theta_i$ that are independently sampled from a uniform distribution, specifically $\theta_i \sim \mathcal{U}(-\pi, \pi)$ \cite{plate1995holographic}. The corresponding HRR atomic vector, $\boldsymbol{x}$, is then obtained by applying the Inverse Fast Fourier Transform (IFFT) to $\boldsymbol{\theta}$:
\begin{equation}
    \label{FHRRtoHRR}
    \boldsymbol{x}=\mathcal{F} ^{-1}\left( e^{j\boldsymbol{\theta }} \right) 
\end{equation}
Here, $\mathcal{F}$ and $\mathcal{F}^{-1}(\cdot)$ represent the Fast Fourier Transform (FFT) and Inverse FFT (IFFT), respectively. When the dimension $d$ is sufficiently large, these randomly generated vectors exhibit pseudo-orthogonality, making them suitable for representing distinct symbols or concepts.

The similarity between any two vectors is a crucial metric for evaluating the distributed representations in VSAs. In FHRR and HRR, cosine similarity is employed to measure the similarity between two vectors \cite{kleyko2022survey}:
\begin{equation}
    \label{cossim} 
    \begin{split}
    sim(\boldsymbol{\theta },\boldsymbol{\phi })&=\frac{1}{d}\sum_{i=1}^d{\cos \left( \theta _i-\varphi _i \right)}
    \\
    sim(\boldsymbol{x},\boldsymbol{y})&=\frac{\boldsymbol{x}\cdot \boldsymbol{y}}{\left| \boldsymbol{x} \right|\left| \boldsymbol{y} \right|}
    \end{split}
\end{equation}
where $\boldsymbol{\theta }$ and $\boldsymbol{\phi }$ denote two FHRR vectors, and $\boldsymbol{x}$ and $\boldsymbol{y}$ two HRR vectors. The similarity $sim(\cdot ,\cdot )$ ranges from -1 to +1, and above two similarity measures are equivalent. The pseudo-orthogonality refers to the case where the similarity $sim(\cdot ,\cdot ) \approx 0$.

\subsection{Basic operations and structured symbolic representations}

\noindent All computations within VSAs are composed of several basic vector algebraic operations, with the primary ones being binding ($\circ$), bundling ($+$) and unbinding ($\oslash$) (Table \ref{Table:BasicOperation}). The binding operation ($\circ$) is employed to form a representation of an object that contains information about the context in which it was encountered \cite{kleyko2022survey}. The bundling operation ($+$), also known as superposition, generates a composite high dimensional vector that combines several lower-level representations. In calculation, binding has a higher priority than bundling. The unbinding operation ($\oslash$), which is the inverse of binding, extracts a constituent from the compound data structure. Binding and bundling are referred to as composition operations, while unbinding is considered a decomposition operation. All operations do not change the vector dimensionality.

Through the combination of these operations, VSAs can effectively achieve structured symbolic representations \cite{kleyko2022survey}. For instance, consider a scene $\boldsymbol{s}$ in which a triangle $\boldsymbol{t}$ is positioned on the left $\boldsymbol{p}_L$ and a circle $\boldsymbol{c}$ on the right $\boldsymbol{p}_R$. This scene can be represented as $\boldsymbol{s} = \boldsymbol{p}_L \circ \boldsymbol{t} + \boldsymbol{p}_R \circ \boldsymbol{c}$ by the role-filler pair \cite{kanerva2009hyperdimensional}. By applying the inverse vector of the left position $\boldsymbol{p}_{L}^{-1}$ to unbind $\boldsymbol{s}$, we can retrieve an approximate vector representing the content at the left position, i.e., $\boldsymbol{p}_{L}^{-1} \otimes \boldsymbol{s} \approx \boldsymbol{t}$. Moreover, the triangle $\boldsymbol{t}$ can itself be a compositional scene, where attributes such as color and size are combined into a triangle scene in a similar manner. This decomposable, structure-sensitive, high-dimensional distributed representation has the potential to disentangle complex scenes while maintaining the advantages of traditional connectionist approaches \cite{hersche2023neuro}.

\subsection{The fractional power encoding method}

\noindent In this study, the rules in RPM are primarily numerical. We introduce the VSA representation of numerical values using the fractional power encoding method (FPE-VSA) \cite{plate1994distributed,Frady_Kleyko_Kymn_Olshausen_Sommer_2021}. Let $x \in \mathbb{R}$ be a real number and $X \in \mathbb{R}^d$ a randomly sampled base vector. The VSA representation $\boldsymbol{v}(x) \in \mathbb{R}^d$ for any value $x$ is obtained by repeatedly binding the base vector $X$ with itself $x$ times, as follows:
\begin{equation}
	\boldsymbol{v}\left( x \right) :=\left( X \right) ^{\left( \circ x \right)} 
	\label{EQU:FPE}
\end{equation}

The FPE method maps arbitrary real numbers to corresponding HD vector, and has the following properties:
\begin{equation}
	\label{FPEsum}
	\boldsymbol{v}\left( x_1+x_2 \right) =\boldsymbol{v}\left( x_1 \right) \circ \boldsymbol{v}\left( x_2 \right) 
\end{equation}
This demonstrates that addition $+$ in the real number domain can be represented by the binding operation $\circ $ in the vector domain.


\begin{table}[t]
    \caption{Basic oprations of FHRR and HRR.}
    \label{Table:BasicOperation}
    \renewcommand{\arraystretch}{1.25}
    \centering
    \begin{tabular}{lll}
        \toprule
        Operations & Impl. on FHRR & Impl. on HRR \\ 
        \midrule
        Binding($\boldsymbol{x } \circ \boldsymbol{y}$) 
        & $\left( \boldsymbol{\theta }+\boldsymbol{\phi} \right) mod\,\,2\pi  $ 
        & $\mathcal{F} ^{-1}\left( \mathcal{F} \left( \boldsymbol{x} \right) \cdot \mathcal{F} \left( \boldsymbol{y} \right) \right)$ \\ 
        Bundling($\boldsymbol{x } + \boldsymbol{y}$) 
        & $angle\left( e^{j\boldsymbol{\theta }}+e^{j\boldsymbol{\phi }} \right) $ 
        & $\boldsymbol{x}+\boldsymbol{y}$ \\ 
        Inverse($\boldsymbol{x}^{-1}$) 
        & $\left( -\boldsymbol{\theta } \right) \,\,mod\,\,2\pi $ 
        & $\mathcal{F} ^{-1}\left( 1/\mathcal{F} \left( \boldsymbol{x} \right) \right) $ \\ 
        Unbinding($\boldsymbol{x } \oslash \boldsymbol{y}$) 
        & $\boldsymbol{\theta^{-1} } \circ \boldsymbol{\phi}$
        & $\boldsymbol{x^{-1} } \otimes \boldsymbol{y}$
        \\
        \bottomrule
    \end{tabular}
\end{table}

\section{Methodology}

\subsection{Atomic HD vectors with semantic representations}

\begin{figure}[t]
	\centering
	\includegraphics[width=1\linewidth]{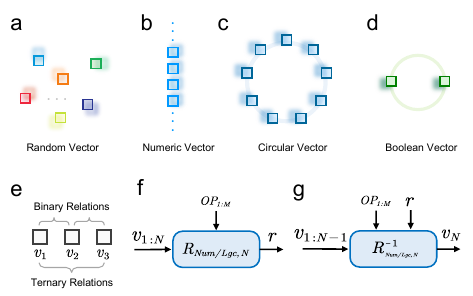} 
	\caption{\textbf{Atomic HD representations and relation functions.} 
		\textbf{(a-d)} The Rel-SAR model utilizes four types of atomic HD vectors. Random Vectors (RVs), sampled independently, are used to represent distinct and unrelated symbols or concepts. Numeric Vectors (NVs) are used to represent real numbers and support VSA-based addition-type arithmetic operations. Circular Vectors (CVs) represent periodic values and enable addition-type arithmetic operations with periodicity. Boolean Vectors (BVs), representing logical values of \textit{False} and \textit{True}, support VSA operations for logical reasoning. \textbf{(e)} In the RAVEN dataset, for a given attribute, the HD attribute representations in a row of three image panels involve binary or ternary relations. \textbf{(f)} Relation functions describe the numerical or logical relations between multiple HD vector representations $\boldsymbol{v}_{1:N}$, where $N=2$ for binary and $N=3$ for ternary relations. These relations are governed by the operator powers $OP_{1:M}$ and the output $\boldsymbol{r}$. \textbf{(g)} For a given relation defined by $OP_{1:M}$ and $\boldsymbol{r}$, inverse relation functions infer the last HD vector representation $\boldsymbol{v}_N$ according to the first $N-1$ representations $\boldsymbol{v}_{1:N-1}$. }
	
	\label{fig:AtomicRepresentation}
\end{figure}

\noindent In neuro-vector-symbolic systems, atomic HD vector representations with meaningful semantics are essential for perception and reasoning. We introduce four types of atomic HD vectors used in our model (Figure \ref{fig:AtomicRepresentation}): Random Vectors (RVs), Numeric Vectors (NVs), Circular Vectors (CVs), and Boolean Vectors (BVs). The definitions and properties of these vectors are universal within the VSA framework.

\subsubsection{\textbf{Random Vector}}
RVs are sampled from specific distributions according to the VSA models, as mentioned in the preliminary section. Due to the absence of numerical or logical relations among RVs and their pseudo-orthogonality in the HD vector space (Figure \ref{fig:AtomicRepresentation}a), they are often used to represent symbols and concepts assumed to be independent and dissimilar. 


\subsubsection{\textbf{Numeric Vector}}
NVs, generated using the fractional power encoding (FPE-VSA, Equation \ref{EQU:FPE}) \cite{plate1994distributed}, are employed to represent real numbers (Figure \ref{fig:AtomicRepresentation}b). NVs $\boldsymbol{v}(r) \in \mathbb{R}^d$ can be used to perform addition-type arithmetic operations through the binding (Equation \ref{FPEsum}) \cite{Frady_Kleyko_Kymn_Olshausen_Sommer_2021}. 

\subsubsection{\textbf{Circular Vector}}
CVs are a special class of NVs used to represent periodic values (Figure \ref{fig:AtomicRepresentation}c). Given a base vector $P$, where each phase of its elements $\rho_i$ is sampled from a discrete distribution ( e.g., for FHRR, $\rho_i\sim \mathcal{U} \left( 2\pi j/L,\forall j\in \left\{ 1,\cdots ,L \right\} \right) $, with $L$ being an even number), CVs are defined as $\boldsymbol{p}(r):=(P)^{(\circ r)}$. These CVs are pseudo-orthogonal to one another and exhibit periodicity with a period of $L$ \cite{Frady_Kleyko_Kymn_Olshausen_Sommer_2021}:
\begin{equation}
	\label{simvalue}
	\boldsymbol{p}\left( r+L \right) =\boldsymbol{p}\left( r \right) 
\end{equation}
If $L$ is odd, the corresponding CVs with period $L$ can be obtained by selecting every other CV from those with period $2L$.

\subsubsection{\textbf{Boolean Vector}}
BVs are a specific type of CVs with a period of $L=2$, used to represent Boolean values (Figure \ref{fig:AtomicRepresentation}d). Following a similar generation method as for CVs, we can generate vectors with a period of $L=2$, $\boldsymbol{e}\left( 0 \right)$ and $\boldsymbol{e}\left( 1 \right)$, to represent \textit{False} and \textit{True}, respectively. Basic logic operations using BVs are implemented as shown in Table \ref{logicoperation}, where $\boldsymbol{a}, \boldsymbol{b} \in \{\boldsymbol{e}(0), \boldsymbol{e}(1)\}$ represent arbitrary Boolean values.

\begin{table}[t]
    \caption{Logic operations implemented by BV}
    \renewcommand{\arraystretch}{1.25}
    \setlength{\tabcolsep}{2.0mm}
    \label{logicoperation}
    \centering
    \begin{tabular}{ll}
        \toprule
         Operation & Implementation on BV \\
        \midrule
        NOT &  $\lnot \boldsymbol{a}=\boldsymbol{a}\circ \boldsymbol{e}\left( 1 \right)  $ \\
        XOR &  $ \boldsymbol{a} \oplus \boldsymbol{b} =\boldsymbol{a}\circ \boldsymbol{b}  $ \\
        AND &  $\boldsymbol{a}\land \boldsymbol{b}= \boldsymbol{a}^{\left( \circ sim\left( \boldsymbol{a},\boldsymbol{b} \right) \right)}$\\
        OR  & $\begin{aligned}
        	\boldsymbol{a}\lor \boldsymbol{b}&=\left( \boldsymbol{a}\oplus \boldsymbol{b} \right) \circ \left( \boldsymbol{a}\land \boldsymbol{b} \right)\\
        	\,\,           &=\boldsymbol{a}\circ \boldsymbol{b}\circ \boldsymbol{a}^{\left( \circ sim\left( \boldsymbol{a},\boldsymbol{b} \right) \right)}\\
        \end{aligned}$ \\
        \bottomrule
    \end{tabular}
\end{table}

\subsection{Relation functions based on atomic HD representations}
\noindent The rules for abductive reasoning in RPM involve binary and ternary relations among the attributes of corresponding objects in each row of three panels (Figure \ref{fig:AtomicRepresentation}e and Figure \ref{fig:RAVENexample}a), as well as numerical and logical relations. In this work, we design general relation functions based on VSA algebra, utilizing the aforementioned atomic vector representations, to be used for rule abductions.

\subsubsection{\textbf{Relation functions}}
Relation functions, which describe the relations between multiple HD vector representations, are categorized into two types: numerical and logical. Among the atomic HD representations, Numeric Vectors (NVs) and Circular Vectors (CVs) are involved in numerical relations, while Boolean Vectors (BVs) are involved in logical relations. 

\textbf{The numerical relation function}, $R_{Num}$, is defined as follows (Figure \ref{fig:AtomicRepresentation}f):
\begin{equation}
	\label{EQU:relationfunction}
	\boldsymbol{r}_{Num}=R_{Num}\left( \boldsymbol{v}_{1:N},OP_{1:M} \right) =\circ _{i=1}^{N}\boldsymbol{v}_{i}^{\left( \circ op_i \right)}
\end{equation}
where $N$ represents the arity of the relation function, and $\boldsymbol{v}_{1:N}:=\left\{ \boldsymbol{v}_i \right\} _{i=1}^{N}$ denotes the input set of HD vector representations. $M$ is the number of operator powers and $OP_{1:M}:=\left\{ op_i \right\} _{i=1}^{M}$ represents the operator powers, which can be considered as parameters of the relation function. The notation $\circ _{i=1}^{N}$ denotes the sequential binding operation applied to the $N$ HD vector representations. $\boldsymbol{r}_{Num}$ is the output HD representation. For the binary numerical relation function, $N=2$ and $M=2$, while for the ternary numerical relation function, $N=3$ and $M=3$. Based on the arithmetic properties of NVs and CVs, $R_{Num}$ can describe the additive relations of these two types of HD vector representations. The combination of $OP_{1:M}$ and $\boldsymbol{r}_{Num}$ determines the specific numerical relation in this vector-symbolic method.


Similarly, \textbf{the simplified logical relation function}, $R_{Lgc}$, is defined as follows (Figure \ref{fig:AtomicRepresentation}f):
\begin{equation}
	\label{EQU:logicrelationfunction}
	\boldsymbol{r}_{Lgc}=R_{Lgc}\left( \boldsymbol{v}_{1:N},OP_{1:M} \right) =\left( op_1\boldsymbol{v}_1\land op_2\boldsymbol{v}_2 \right) \circ op_3\boldsymbol{v}_3
\end{equation}
where $\boldsymbol{v}_{1:N}:=\left\{ \boldsymbol{v}_i \right\} _{i=1}^{N} \in \{\boldsymbol{e}(0), \boldsymbol{e}(1)\}$ denotes the input set of BVs. The full version of logical relation function is described in Appendix \ref{sec:appendix_1}. Here, we consider only the ternary logical relation, so $N=3$ and $M=3$. The parameter $OP_{1:M}:=\left\{ op_i \right\} _{i=1}^{M}$, where $op_i\in \{0,1\}$ determines whether to negate $\boldsymbol{v}_i$, with negation ($\lnot$) applied when $op_i=1$ and no negation applied when $op_i=0$ (see Appendix \ref{sec:appendix_1}). The symbol $\land$ denotes the AND operation, as shown in Table \ref{logicoperation}. Based on the computational properties of BVs detailed in Table \ref{logicoperation}, $R_{Lgc}$ can describe the ternary logical relations involved in RPM. The combination of the operator $OP_{1:M}$ and the output $\boldsymbol{r}_{Lgc}$ determines the specific logical relation in this vector-symbolic method. 

\begin{figure*}[t]
	\centering
	\includegraphics[width=0.60\linewidth]{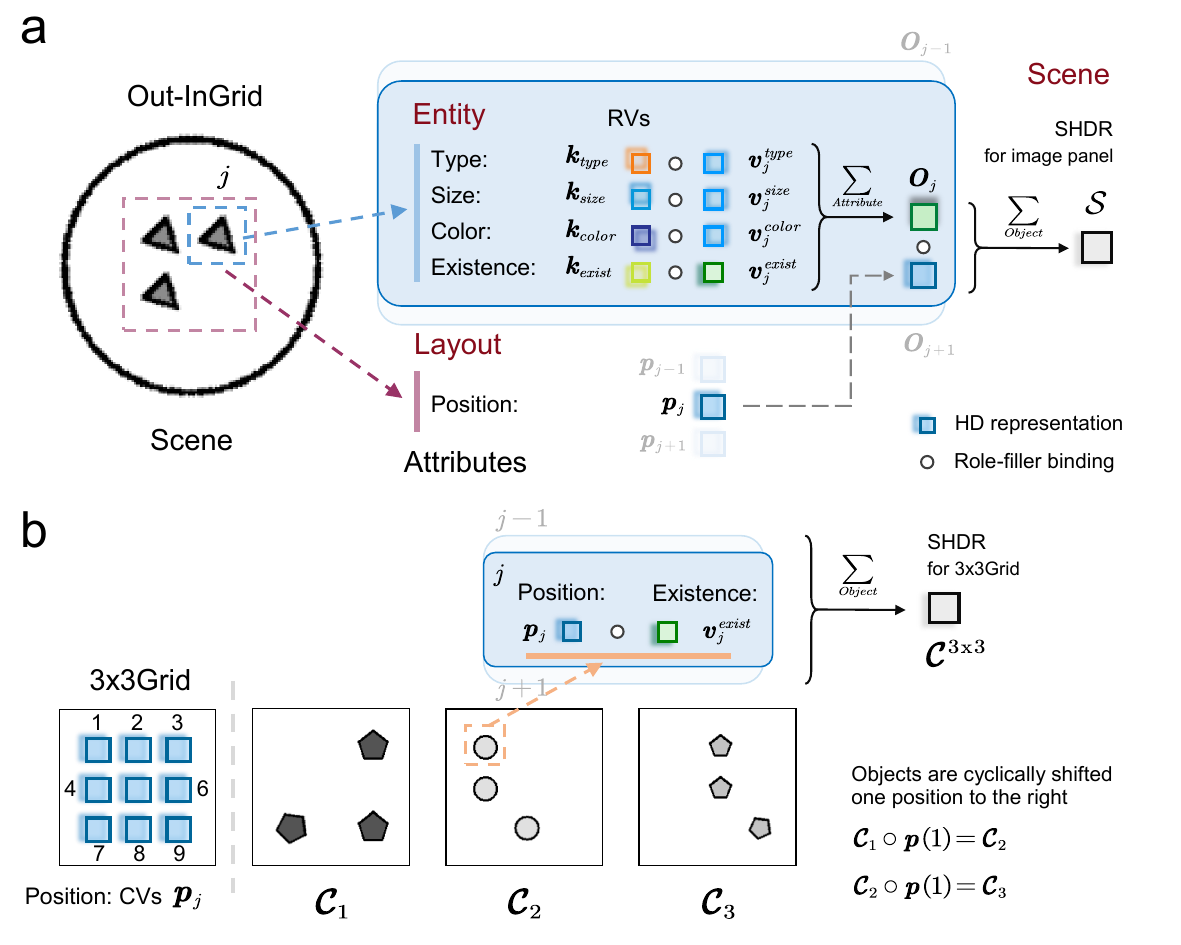} 
	\caption{\textbf{Structured HD representations (SHDR) for the image panel and the \textsf{nxn Grid}.} 
		\textbf{(a)} SHDR for the image panel (Equation \ref{SHDR}). Taking the \textsf{Out-InGrid} configuration as an example, an image panel contains multiple objects, each with four entity attributes: \textit{type}, \textit{size}, \textit{color}, and \textit{existence}. Through the first layer of role-filler binding, these attributes are combined to form a SHDR for each object. Additionally, at the layout level, each object has a \textit{position} attribute. By applying a second layer of role-filler binding, the SHDR for the entire image panel is constructed. \textbf{(b)} SHDR for the \textsf{nxn Grid} (Equation \ref{SHDR_nxn}). Taking the \textsf{3x3Grid} configuration as an example, the position vectors for all objects are represented using circular vectors (CVs) with a period of $3 \times 3 = 9$. The SHDR $\mathcal{C}^{3\mathrm{x}3}$ for this \textsf{3x3Grid} is obtained by performing role-filler binding between the corresponding \textit{position} vectors $\boldsymbol{p}_j$ and \textit{existence} vectors $\boldsymbol{v}_j$. Due to the periodic nature of the \textit{position} vectors, as all objects shift positions cyclically, the SHDR undergoes a binding operation with the \textit{position} vectors corresponding to the magnitude of the shift.
		}
	
	\label{fig:SHDR}
\end{figure*}

\subsubsection{\textbf{Inverse relation functions}}
Rule execution in RPM requires inferring the third attribute value based on the first two attribute values in a row of panels, given a known relation. It represents an inverse problem of rule abduction. In the vector-symbolic method, given the operator power ${OP_{1:M}}$ and the output $\boldsymbol{r}$, the last vector representation $\boldsymbol{v}_N$ can be inferred from the first $N-1$ inputs $\boldsymbol{v}_{1:N-1}$ using the inverse of the relation functions (Figure \ref{fig:AtomicRepresentation}g). According to Equation \ref{EQU:relationfunction}, the inverse numerical relation function is defined as follows:
\begin{equation}
\begin{aligned}
	\boldsymbol{v}_N&=R_{Num}^{-1}\left( \boldsymbol{v}_{1:N-1},OP_{1:M},\boldsymbol{r} \right)\\
	&=\left( \circ _{i=1}^{N-1}\boldsymbol{v}_{i}^{\left( \circ \left( -op_i/op_M \right) \right)} \right) \circ \boldsymbol{r}^{\left( \circ \left( -1/op_M \right) \right)}\\
\end{aligned}
\label{EQU:Inverserelationfunction}
\end{equation}

Similarly, according to Equation \ref{EQU:logicrelationfunction}, the inverse logical relation function is defined as follows:
\begin{equation}
	\begin{aligned}
		\boldsymbol{v}_N&=R_{Lgc}^{-1}\left( \boldsymbol{v}_{1:N-1},OP_{1:M},\boldsymbol{r} \right)\\
		&=op_3\left( op_1\boldsymbol{v}_1\land op_2\boldsymbol{v}_2 \right)\\
	\end{aligned}
\label{EQU:Inverselogicrelationfunction}
\end{equation}.

\subsection{Structured high-dimensional representation and its attribution decomposition}

\noindent VSA can create structured symbolic representations using atomic HD vector representations and decouple them directly from these structures through algebraic operations \cite{hersche2023neuro}. This subsection presents the process of constructing a structured HD representation (SHDR) for an image panel and its decomposition to retrieve individual attribute representations. Additionally, an SHDR for the \textsf{nxn Grid} ($n=2,3$) at the component level is also introduced.

\subsubsection{\textbf{SHDR for the image panel}}
In RAVEN dataset, each image panel consists of objects, with each object characterized by multiple attributes. Consequently, the structured HD representation (SHDR) for each image panel can be obtained through two layers of role-filler bindings (Figure \ref{fig:SHDR}a). First, the bundling operation is used to construct an SHDR for each object at the entity level by combining its attributes. Then, another bundling operation aggregates these object-level representations to construct a SHDR of the image panel at the scene level. Therefore, each image panel $\mathcal{X} \in \mathbb{R} ^{r\times r}$, with a resolution $r \times r$, can be represented by an SHDR $\mathcal{S} \in \mathbb{R}^{d}$ as follows:
\begin{equation}
	\begin{aligned}
		\label{SHDR} 
		\mathcal{S} &=\sum_{j=1}^{N_{pos}}{\boldsymbol{p}_j\circ \boldsymbol{O}_j} \\
		&=\sum_{j=1}^{N_{pos}}{\boldsymbol{p}_j\circ \left( \sum_{attr\in ATTR}{\boldsymbol{k}_{attr}\circ \boldsymbol{v}_{j}^{attr}} \right)}\\
	\end{aligned}
\end{equation}
Here, $\boldsymbol{O}_j$ represents the SHDR of the $j$th object with different attributes at the entity level, incorporating attributes such as type, size, color, and existence.  The attribute set is $ATTR = \left\{ type,size,color,exist \right\}$. At the entity level, the key vector $\boldsymbol{k}_{attr}$ denotes the class of a specific attribute $attr \in ATTR$, while the value vector $\boldsymbol{v}_{j}^{attr}$ indicates the attribute's value at the position $j$. At the scene level, the position vector $\boldsymbol{p}_j$ specifies the location of the $j$-th object. 

\subsubsection{\textbf{Representation decomposition}}
Given an estimated SHDR $\hat{\mathcal{S}} \in \mathbb{R}^{d}$ of an image panel, all SHDRs of objects $\hat{\boldsymbol{O}}_j$ at the entity level, along with the corresponding attribute representations $\hat{\boldsymbol{v}}_{j}^{attr}$, can be derived through a series of unbinding operations \cite{kleyko2022survey}. The decomposition process is shown as follows:
\begin{equation}
\begin{cases}
	\hat{\boldsymbol{O}}_j=\boldsymbol{p}_j \oslash \hat{\mathcal{S}} =\boldsymbol{p}_{l}^{-1}\circ \hat{\mathcal{S}}\\
	\hat{\boldsymbol{v}}_{j}^{attr}=\boldsymbol{k}_{attr} \oslash \hat{\boldsymbol{O}}_j=\boldsymbol{k}_{attr}^{-1}\circ \hat{\boldsymbol{O}}_j\\
\end{cases}
\label{SHDR_decomposition}
\end{equation}
It is important to note that due to inaccuracies in the estimated SHDR  $\hat{\mathcal{S}}$ and the noise introduced by the unbinding operation, the estimated attribute representations $\hat{\boldsymbol{v}}_{j}^{attr}$ may not fully match the original $\boldsymbol{v}_{j}^{attr}$ used in Equation \ref{SHDR}. 


\subsubsection{\textbf{SHDR for the \textsf{nxn Grid} component}}
In the RAVEN dataset, three figure configurations—\textsf{2x2Grid}, \textsf{3x3Grid}, and \textsf{Out-InGrid}—include components where objects are arranged in an \textsf{nxn grid} pattern at the layout level \cite{zhang2019raven}. Since the positions in the \textsf{nxn Grid} involve component-level rule reasoning, the SHDR for the \textsf{nxn Grid} component ($n=2,3$), focusing only on positions and object existence, is introduced as follows (Figure \ref{fig:SHDR}b):
\begin{equation}
\mathcal{C} ^{\mathrm{nxn}}=\sum_{j=1}^{n\times n}{\boldsymbol{p}_j\circ \boldsymbol{v}_{j}^{exist}}
\label{SHDR_nxn}
\end{equation}



\subsection{Rules from the perspective of relation functions}

\begin{table*}[ht]
	\centering
	\caption{Attribute representations and the relation functions involved in rule abductions}
	\label{Tab:AttributeRepresentationRule}
	
	\begin{tblr}{
			hlines = {},
			vlines = {},
			colspec={ccccccc},
			cells={mode=text},
			cell{1}{3} = {c = 2}{halign = c},
			cell{1}{5} = {c = 2}{halign = c},
			cell{2}{1} = {r = 3}{valign = m},
			cell{5}{1} = {r = 5}{valign = m},
			cell{6}{2} = {r = 4}{valign = m},
			cell{2}{3} = {r = 4}{valign = m},
			cell{6}{3} = {r = 3}{valign = m},
			cell{2}{4} = {r = 4}{valign = m},
			cell{6}{4} = {r = 3}{valign = m},
			cell{2}{6} = {r = 7}{valign = m},
			cell{2}{7} = {r = 2}{valign = m},
			cell{4}{7} = {r = 2}{valign = m},
			cell{6}{7} = {r = 2}{valign = m}
		}
		
		Level & Attributes & HD representations &  & Rules in RAVEN and their types&  & Relation functions \\
		
		Entity & Type & $\begin{array}{c}
			\boldsymbol{v}^{type}\\
			\boldsymbol{v}^{size}\\
			\boldsymbol{v}^{color}\\
			\boldsymbol{v}^{num}\\
		\end{array}$ & Atomic (NVs) & Constant & Numerical rules & Binary + Numerical \\
		
		& Size &   & & Progression & & \\
		
		& Color &   & & Distribute Three & & Ternary + Numerical \\
		
		Layout & Number &  & & Arithmetic & & \\
		
		& Position &  $\mathcal{C} ^{\mathrm{nxn}}=\sum_{j=1}^{n\times n}{\boldsymbol{p}_j\circ \boldsymbol{v}_{j}^{exist}}$ & $\begin{array}{c}
			\mathrm{SHDR}\\
			\boldsymbol{p}_j: \mathrm{CVs}\\
			\boldsymbol{v}_{j}^{exist}: \mathrm{RVs}\\
		\end{array}$ & Constant & & Binary + Numerical \\
		
		& & & & Progression & &  \\
		
		& & & & Distribute Three & & Ternary + Numerical\\
		
		& & $\boldsymbol{v}^{exist}$ & Atomic (BVs) & Arithmetic & Logical rules & Ternary + logical \\
		
	\end{tblr}
\end{table*}

\begin{table*}[ht]
	\centering
	\caption{Rules and Corresponding combinations of $OP_{1:M}$ and $r_{Num}$ in Relation functions}
	\label{Tab:RulesAndOP}
	
	\begin{tblr}{
			hlines = {},
			vlines = {},
			colspec={cccccccc},
			cells={mode=text},
			cell{1}{1} = {r = 2, c = 2}{halign = c, valign = m},
			cell{1}{7} = {r = 2}{valign = m},
			cell{1}{8} = {r = 2}{valign = m},
			cell{3}{1} = {r = 6}{valign = m},
			cell{3}{2} = {c = 2}{halign = l},
			cell{4}{2} = {r = 2}{valign = m,halign = l},
			cell{6}{2} = {r = 2}{valign = m,halign = l},
			cell{8}{2} = {c = 2}{halign = c,halign = l},
			cell{9}{1} = {r = 2}{valign = m},
			cell{9}{2} = {r = 2}{valign = m,halign = l}
		}
		
		Rules in RAVEN & & Ternary & $op_{1}$ & $op_{2}$ & $op_{3}$ & $r$ & Examples (Rule $\rightarrow$ Relation function) \\
		  & & Binary &  &  $op_{1}$ &  $op_{2}$ &  &  \\
		  
		Numerical rules & Constant & & 0 & $-1$ & $+1$ & 0 & $v_1=v_2\rightarrow \boldsymbol{v}\left( 0 \right) =\boldsymbol{v}_{1}^{\left( \circ \left( -1 \right) \right)}\circ \boldsymbol{v}_{2}^{\left( \circ 1 \right)}$ \\
		
		 & Progression & $+$ & 0 & $-1$ & $+1$ & $+1$,$+2$ & $v_1+1=v_2\rightarrow \boldsymbol{v}\left( +1 \right) =\boldsymbol{v}_{1}^{\left( \circ \left( -1 \right) \right)}\circ \boldsymbol{v}_{2}^{\left( \circ 1 \right)}$	\\
		 &             & $-$ & 0 & $-1$ & $+1$ & $-1$,$-2$ & $v_1-2=v_2\rightarrow \boldsymbol{v}\left( -2 \right) =\boldsymbol{v}_{1}^{\left( \circ \left( -1 \right) \right)}\circ \boldsymbol{v}_{2}^{\left( \circ 1 \right)}$	\\
		 
		 & Arithmetic & $+$ & $-1$ & $-1$ & $+1$ & $0$ & $v_1+v_2=v_3\rightarrow \boldsymbol{v}\left( 0 \right) =\boldsymbol{v}_{1}^{\left( \circ \left( -1 \right) \right)}\circ \boldsymbol{v}_{2}^{\left( \circ (-1) \right)} \circ \boldsymbol{v}_{3}^{\left( \circ 1  \right)}$	\\
		 
		 &            & $-$ & $-1$ & $+1$ & $+1$ & $0$ & $v_1-v_2=v_3\rightarrow \boldsymbol{v}\left( 0 \right) =\boldsymbol{v}_{1}^{\left( \circ \left( -1 \right) \right)}\circ \boldsymbol{v}_{2}^{\left( \circ 1 \right)} \circ \boldsymbol{v}_{3}^{\left( \circ  1  \right)}$	\\
		 
		 & Distribute Three &  & $+1$ & $+1$ & $+1$ & \textit{Any} & $v_1+v_2+v_3=Any\rightarrow \boldsymbol{Any} =\boldsymbol{v}_{1}^{\left( \circ 1 \right)}\circ \boldsymbol{v}_{2}^{\left( \circ 1 \right)} \circ \boldsymbol{v}_{3}^{\left( \circ 1  \right)}$	\\
		 
		 Logical rules & Arithmetic & $+$ & $+1$ & $+1$ & $+1$ & 0 & $\boldsymbol{e}\left( 0 \right) =\left( \boldsymbol{e}\left( 1 \right) \circ \boldsymbol{v}_1\land \boldsymbol{e}\left( 1 \right) \circ \boldsymbol{v}_2 \right) \circ \left( \boldsymbol{e}\left( 1 \right) \circ \boldsymbol{v}_3 \right) $	\\
		 
		  &  & $-$ & $0$ & $+1$ & $0$ & 0 & $\boldsymbol{e}\left( 0 \right) =\left( \boldsymbol{e}\left( 0 \right) \circ \boldsymbol{v}_1\land \boldsymbol{e}\left( 1 \right) \circ \boldsymbol{v}_2 \right) \circ \left( \boldsymbol{e}\left( 0 \right) \circ \boldsymbol{v}_3 \right) $	\\
	\end{tblr}
\end{table*}

\noindent The RAVEN dataset contains $4$ rules—\textsl{Constant}, \textsl{Progression}, \textsl{Arithmetic}, and \textsl{Distribute Three}—which operate on $5$ rule-governing attributes \cite{zhang2019raven}. These $5$ attributes include $3$ entity-level attributes: \textit{Type}, \textit{Size}, and \textit{Color}, as well as $2$ layout-level attributes: \textit{Number} and \textit{Position}. In this study, the HD representations of these attribute values during rule reasoning and relations between rules and relation functions are shown in Table \ref{Tab:AttributeRepresentationRule}. 

For the attributes \textit{Type}, \textit{Size}, \textit{Color}, and \textit{Number}, the four involved rules follow additive arithmetic operations, meaning the attribute values $\boldsymbol{v}^{attr}$ ($attr \in \left\{ type,size,color, number \right\}$) are represented using Numeric Vectors (NVs). Therefore, these rules can be defined using the numerical relation function (Equation \ref{EQU:relationfunction}): \textsl{Constant} and \textsl{Progression} correspond to binary relation functions, while \textsl{Arithmetic} and \textsl{Distribute Three} correspond to ternary relation functions. Each rule is associated with specific combinations of $OP_{1:M}$ and $\boldsymbol{r}_{Num}$, and corresponding details are shown in Table \ref{Tab:RulesAndOP}.

For the attribute \textit{Position}, the rules \textsl{Constant}, \textsl{Progression}, and \textsl{Distribute Three} primarily refer to an \textsf{nxn Grid} with multiple objects, which, in an overall sense, follow additive arithmetic operations. Therefore, we use the SHDR $\mathcal{C} ^{\mathrm{nxn}}$ for the \textsf{nxn Grid} (Equation \ref{SHDR_nxn}) to represent the attributes required by these three rules. Since \textsl{Progression} involves a cyclic left or right shift of all objects (Figure \ref{fig:SHDR}b), the position vectors $\boldsymbol{p}_j$ in set $\mathcal{C} ^{\mathrm{nxn}}$ during rule reasoning are represented by Circular Vectors (CVs). The object existence vectors $\boldsymbol{v}_{j}^{exist}$ are represented by Random Vectors (RVs). These three rules can also be described using numerical relation functions (Equation \ref{EQU:relationfunction}). Take the rule \textsl{Progression} (+1) as an example, where the positions of objects undergo a cyclic right shift. In Figure \ref{fig:SHDR}b, the SHDRs of the \textsf{3x3 Grid} across a row of three panels exhibit two numerical relations:$\mathcal{C} _1\circ \boldsymbol{p}\left( 1 \right) =\mathcal{C} _2$ and $\mathcal{C} _2\circ \boldsymbol{p}\left( 1 \right) =\mathcal{C} _3$, which can be defined using a binary numerical relation function.

In addition, the rule \textsl{Arithmetic} on the attribute \textit{Position} is belong to the logical rule \cite{zhang2019raven}. The attribute values $\boldsymbol{v}_{j}^{exist}$ are also represented using Boolean Vectors (BVs) that can be operated as shown in Table \ref{logicoperation}. Therefore, the rule \textsl{Arithmetic} for \textit{Position} corresponds to the ternary logical relation function (Equation \ref{EQU:logicrelationfunction}), and corresponding details about the combinations of $OP_{1:M}$ and $\boldsymbol{r}_{Num}$ in the relation function are shown in Table \ref{Tab:RulesAndOP}.

\subsection{The Systematic Abductive Reasoning model}
\begin{figure*}[t]
	\centering
	\includegraphics[width=0.71\linewidth]{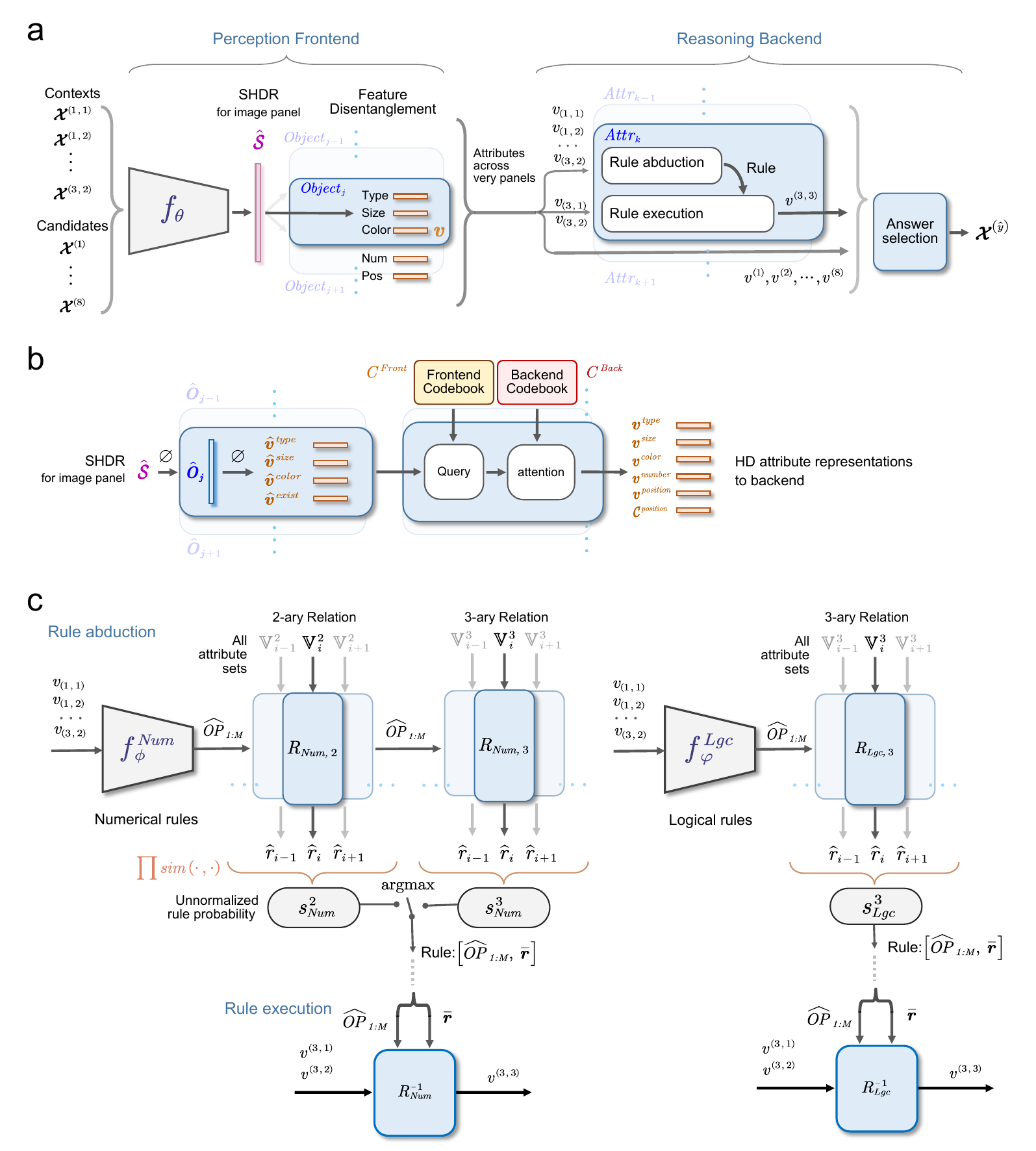} 
	\caption{\textbf{The Systematic Abductive Reasoning model with diverse relation representations (Rel-SAR).} 
		\textbf{(a)} Overall architecture of Rel-SAR model. Our model consists of a visual perception frontend, which processes object attributes for $8$ context and $8$ candidate image panels in a RPM test, and a reasoning backend that performs symbolic arithmetic and logical reasoning. The perception frontend utilizes a neural network, $f_{\theta}$, to obtain the SHDR of each image panel, and then perceives attributes in the form of HD representations required by the downstream reasoning. In the reasoning backend, the rule abduction module extracts rules for each attribute representation using relation functions. The rule execution module then predicts the missing panel's attribute representations based on inverse relation functions. Finally, the answer selection module compares the predicted attributes of the missing panel with those in the candidate panels and selects the option with the highest similarity. \textbf{(b)} Given the predicted SHDR for each panel, the SHDR of all objects and their corresponding HD attribute representations can be obtained via representation decomposition. Subsequently, the estimated HD attribute representations are refined in two steps: querying the frontend codebook and applying attention based on the backend codebook. This process produces HD attribute representations suitable for backend reasoning, including attributes such as \textit{type}, \textit{size}, \textit{color}, \textit{number}, and \textit{position}. \textbf{(c)} In the rule abduction module, the rule learners $f_{\phi }^{Num}$ and $f_{\varphi }^{Lgc}$ predict the operator powers $\widehat{OP}_{1:M}$ for numerical and logical relation functions based on attributes in the context panels. These predicted $\widehat{OP}_{1:M}$ ensure that all binary or ternary relation input pairs ($\mathbb{V}^{N}, N=2,3$) produce the same output $\boldsymbol{\hat{r}}$ when processed through their respective relation functions. Therefore, the rule defined by $\widehat{OP}_{1:M}$ and $\boldsymbol{\hat{r}}$ with the highest overall $\boldsymbol{\hat{r}}$ similarity, also viewed as unnormalized probability, is considered the underlying rule. The rule execution module then predicts the attributes of the missing panel using inverse relation functions with the estimated rules.   }
	
	\label{fig:ModelArchitecture}
\end{figure*}

\noindent In this section, we present the Systematic Abductive Reasoning model with diverse relation representations (Rel-SAR), inspired by the NVSA \cite{hersche2023neuro}. An overview of Rel-SAR is depicted in Figure \ref{fig:ModelArchitecture}a. Similar to previous neuro-symbolic models for abstract visual reasoning, Rel-SAR combines a neural visual perception frontend with a symbolic reasoning backend, both utilizing VSA representations with meaningful semantics to facilitate systematic reasoning. The perception frontend employs a neural network to extract the SHDR $\mathcal{S}$ of each image panel $\mathcal{X}$ in the RPM and achieves feature disentanglement from the SHDR using representation decomposition to obtain the HD representations of attributes ($\boldsymbol{v}$, $\boldsymbol{p}$ and $\mathcal{C}$: Table \ref{Tab:AttributeRepresentationRule}) required for reasoning in the backend. The reasoning backend consists of three main modules: the rule abduction module, the rule execution module, and the answer selection module. The rule abduction module extracts the corresponding rules ($OP_{1:M}$ and $\boldsymbol{r}$: Table \ref{Tab:RulesAndOP}) for each attribute representation according to appropriate relation function (Equation \ref{EQU:relationfunction} and \ref{EQU:logicrelationfunction}, Table \ref{Tab:AttributeRepresentationRule}). Subsequently, the rule execution module uses these rules to predict the representations of the missing panel's attributes according to corresponding inverse relation functions (Equation \ref{EQU:Inverserelationfunction} and \ref{EQU:Inverselogicrelationfunction}). Finally, the answer selection module compares the predicted attribute representations of the missing panel with the available options in the answer panels and selects the answer. 

\subsubsection{\textbf{Perception frontend}}
The perception frontend operates independently on each of the 16 image panels to extract the HD representations of attributes required for abductive reasoning (Figure \ref{fig:ModelArchitecture}a and Figure \ref{fig:ModelArchitecture}b). For a given image panel $\mathcal{X}^{ind} \in \mathbb{R} ^{r\times r}$, where $ind\in \left\{ \left( 1,1 \right) ,\left( 1,2 \right) ,\cdots ,\left( 3,2 \right) \right\}$ for 8 contexts and $ind\in \left\{ 1,2,\cdots ,8 \right\}$ for 8 candidates, the frontend uses a trainable neural network (ResNet-50) to map the image panel to its estimated SHDR $\hat{\mathcal{S}}^{ind} \in \mathbb{R} ^d$: $f_{\theta}:\mathcal{X} \rightarrow \hat{\mathcal{S}}$, where $\theta$ represents the trainable parameters of the network. Theoretically, the expected SHDR $\mathcal{S}$ for each panel should be organized from the corresponding attribute representations as described by Equation \ref{SHDR}. Therefore, the learning objective of $f_{\theta}$ is to minimize the difference between its output $\hat{\mathcal{S}}$ and the theoretical SHDR $\mathcal{S}$, formulated as:
\begin{equation}
	\label{PerceptionValueFunction}
	\underset{\theta}{\min}\left\| f_{\theta}\left( \mathcal{X} ;\theta \right) -\mathcal{S} \right\| 
\end{equation}

Subsequently, the estimated SHDR $\hat{\mathcal{S}}^{ind}$ for each panel undergoes representation decomposition (Figure \ref{fig:ModelArchitecture}b), as described in Equation \ref{SHDR_decomposition}, to obtain the estimated HD attribute representations for each object, including \textit{Type} ($\hat{\boldsymbol{v}}^{type}_j$), \textit{size} ($\hat{\boldsymbol{v}}^{size}_j$), \textit{color} ($\hat{\boldsymbol{v}}^{color}_j$), and \textit{existence} ($\hat{\boldsymbol{v}}^{exist}_j$), where $j$ denotes the position index of the corresponding object. 

The HD attribute representations are expected to be selected from a set of frontend codebooks for the available attributes of interested in the RAVEN dataset (Figure \ref{fig:ModelArchitecture}b). These frontend codebooks include $C_{Num}^{Front}:=\left\{ \boldsymbol{v}(r) \right\} _{r=0}^{9}\cup \left\{ \boldsymbol{v}_{null} \right\}$ and $C_{Lgc}^{Front}:=\left\{ \boldsymbol{e}(r) \right\}_{r=0}^{1}$, which represent the numerical value and logic, respectively. $\boldsymbol{v}_{null}$ represents the null attribute representation when there is no object. To improve the neural network's performance in encoding the SHDR of an image panel, all hypervectors in these frontend codebooks are randomly and independently generated as RVs, rather than using NVs, CVs, or BVs. 

However, the estimated HD attribute representations for each object, $\hat{\boldsymbol{v}}^{attr}_j$ ($attr \in \{type, size, color, exist \}$), cannot be directly applied to the reasoning backend. First, these representations contain noise introduced by the bundling operation in the form SHDR. Second, as they are expected to be derived from the frontend codebooks of RVs, there are no intrinsic arithmetic or logical relations between the $\hat{\boldsymbol{v}}^{attr}_j$s, which hinders effective reasoning.

To address these issues, we adopt an approach similar to the attention mechanism \cite{vaswani2017attention} to obtain the HD attribute representations suitable for the reasoning backend (Figure \ref{fig:ModelArchitecture}b). In the query stage, we use the estimated HD attribute representations ($\hat{\boldsymbol{v}}^{attr}_j$) as query vectors to compute their similarity with all possible vectors for the corresponding attributes in the frontend codebooks. In the attention stage, these similarity scores are then used as attention weights to perform a weighted summation of the corresponding vectors from the backend codebooks, in which all hypervectors are generated according to their attribute type as shown in Table \ref{Tab:AttributeRepresentationRule}. The backend codebooks consist of $C_{Num}^{Back}$ (NVs), $C_{Lgc, BV}^{Back}$ (BVs), $C_{Lgc, RV}^{Back}$ (RVs), and $C_{Pos,\mathrm{nxn}}^{Back}:=\left\{ \boldsymbol{p}_r \right\} _{r=1}^{n^2}$ (CVs for positions in \textsf{nxn Grid}). The updated HD attribute representations obtained after the weighted summation can be utilized in the reasoning backend. The details are provided below.  

\textbf{The query stage:} For each attribute $attr\in \left\{ type,size,color \right\} $, we compute the cosine similarity between the estimated HD attribute representation $\hat{\boldsymbol{v}}^{attr}_j$  and all possible vectors of $C_{Num}^{Front}$ in the frontend codebooks.
\begin{equation}
W_{j}^{attr}=sim\left( \hat{\boldsymbol{v}}_{j}^{attr},C_{Num}^{Front} \right) 
\end{equation}
where $W_{j}^{attr} (r)$ ($r \in \{0,1,...,9, null\}$) represents the attention weights corresponding to the value $r$ of attribute $attr$ at the $j$th position, based on the query similarity. Similarly, the attention weights for the attribute $attr\in \left\{ exist \right\} $ can be obtained by querying the logic codebook $C_{Lgc}^{Front}$ as follows:
\begin{equation}
W_{j}^{exist}=softmax\left( \beta \cdot sim\left( \hat{\boldsymbol{v}}_{j}^{exist},C_{Lgc}^{Front} \right) \right) 
\label{query_exist}
\end{equation}
where $W_{j}^{exist} (r)$ ($r \in \{0,1\}$) corresponds to the presence and absence of the object at $j$th position, respectively. Here, we use the $softmax$ function to normalize the weights, and $\beta$ denotes the inverse softmax temperature. 

\textbf{The attention stage:} The HD attribute representations required by the reasoning backend involve the entity-level attributes \textit{Type}, \textit{Size}, and \textit{Color}, as well as layout-level attributes \textit{Number}, \textit{Position} (Table \ref{Tab:AttributeRepresentationRule}). For the numerical attribute $attr\in \left\{ type,size,color \right\} $, the corresponding updated HD representation $\boldsymbol{v}_{j}^{attr}$ can be obtained through the weighted summation on the numerical backend codebook $C_{Num}^{Back}$ as follows: 
\begin{equation}
\boldsymbol{v}_{j}^{attr}=\sum_{r\in \left\{ 0,...,9,null \right\}}^{}{W_{j}^{attr}\left( r \right) \cdot \boldsymbol{v}\left( r \right)},\boldsymbol{v}\left( r \right) \in C_{Num}^{Back}
\end{equation}

For the logical existence attribute $attr\in \left\{ exist \right\}$, its updated HD representation $\boldsymbol{v}_{j}^{exist}$ are obtained through the weighted summation on the backend codebook $C_{Lgc, RV}^{Back}$ and $C_{Lgc, BV}^{Back}$, respectively, as follows:
\begin{equation}
\boldsymbol{v}_{j}^{exist, VT}=\sum_{r\in \left\{ 0,1 \right\}}^{}{W_{j}^{exist}\left( r \right) \cdot \boldsymbol{e}\left( r \right)},\boldsymbol{e}\left( r \right) \in C_{Lgc, VT}^{Back}
\end{equation}
where $VT \in \{RV, BV\}$ represents the type of atomic HD vectors. 

Additionally, we introduce the overall HD attribute representations for the \textit{Type}, \textit{Size}, and \textit{Color} attributes within the \textsf{nxn Grid}. These representations $\boldsymbol{v}_{\mathrm{nxn}}^{attr}$ ($attr\in \left\{ type,size,color \right\} $) can be obtained by bundling corresponding HD attribute representations of all objects in the \textsf{Grid} with their attention weights of \textit{existence} as follows:
\begin{equation}
\boldsymbol{v}_{\mathrm{nxn}}^{attr}=\sum_{j=1}^{n^2}{W_{j}^{exist}\left( 1 \right) \cdot \boldsymbol{v}_{j}^{attr}}
\end{equation}

For the layout-level attribute \textit{Number}, its HD attribute representation $\boldsymbol{v}^{number}$ is obtained by projecting the sum of the attention weights of presence to FPE-VSA as follows:  
\begin{equation}
\boldsymbol{v}^{number}=\boldsymbol{v}^{\left( \otimes \sum_{j=1}^{n^2}{W_{j}^{exist}\left( 1 \right)} \right)}
\end{equation}
where $\boldsymbol{v}$ is the base vector of the numerical backend codebook $C_{Num}^{Back}$.

The layout-level attribute \textit{Position} within the \textsf{nxn Grid} involves both numerical and logical rules (Table \ref{Tab:AttributeRepresentationRule}). Therefore, its HD attribute representations correspond to two distinct rules: the logical representation of each individual object $\boldsymbol{v}_{j}^{position}$ and the overall HD position representation $\mathcal{C}^{position}$ of the entire \textsf{nxn Grid}. The former is an HD existence representation with logical computational properties, that is, $\boldsymbol{v}_{j}^{position}=\boldsymbol{v}_{j}^{exist,BV}$. Inspired from SHDR for the \textsf{nxn Grid} in Equation \ref{SHDR_nxn}, the overall HD position representation $\mathcal{C}^{position}$ can be obtained as follows:
\begin{equation}
\mathcal{C}^{position}_{\mathrm{nxn}}=\sum_{j=1}^{n\times n}{W_{j}^{exist}\left( 1 \right) \cdot \boldsymbol{p}_j\circ \boldsymbol{v}_{j}^{exist,RV}}\,, \boldsymbol{p}_j\in C_{Pos,\mathrm{nxn}}^{Back}
\end{equation}

\subsubsection{\textbf{Reasoning backend}}

The Rel-SAR model efficiently implements systematic abductive reasoning by leveraging HD attribute representations and VSA-based relation functions. HD attribute representations from the frontend are transformed into the HD vector space, enabling VSA operations on both numerical and logical relation functions (Equation \ref{EQU:relationfunction}-\ref{EQU:Inverselogicrelationfunction}). Consequently, the reasoning backend can perform systematic rule abduction and execution based on these relational functions, without requiring extensive use of explicit rule templates.

\begin{table}[t]
	\caption{Attribute sets for n-ary relation.}
	\label{Table:RelationSets}
        \renewcommand{\arraystretch}{1.25}
	\centering
	\begin{tabular}{ll}
		\toprule
		$\mathbb{V} ^2$ & $\left( \boldsymbol{v}_{\left( 1,1 \right)},\boldsymbol{v}_{\left( 1,2 \right)} \right)$,
		$\left( \boldsymbol{v}_{\left( 1,2 \right)},\boldsymbol{v}_{\left( 1,3 \right)} \right)$,
		$\left( \boldsymbol{v}_{\left( 2,1 \right)},\boldsymbol{v}_{\left( 2,2 \right)} \right)$,\\
		& $\left( \boldsymbol{v}_{\left( 2,2 \right)},\boldsymbol{v}_{\left( 2,3 \right)} \right)$,
		$\left( \boldsymbol{v}_{\left( 3,1 \right)},\boldsymbol{v}_{\left( 3,2 \right)} \right)$;
		$\left( \boldsymbol{v}_{\left( 3,2 \right)},\boldsymbol{v}_{\left(y\right)} \right)$ 
		\\
		\midrule
		$\mathbb{V} ^3$ & $\left( \boldsymbol{v}_{\left( 1,1 \right)},\boldsymbol{v}_{\left( 1,2 \right)},\boldsymbol{v}_{\left( 1,3 \right)} \right)$,
		$\left( \boldsymbol{v}_{\left( 2,1 \right)},\boldsymbol{v}_{\left( 2,2 \right)},\boldsymbol{v}_{\left( 2,3 \right)} \right)$; \\ 
		&$\left( \boldsymbol{v}_{\left( 3,1 \right)},\boldsymbol{v}_{\left( 3,2 \right)},\boldsymbol{v}_{\left(y\right)} \right)$ \\ 
		\bottomrule
	\end{tabular}
\end{table}

\textbf{Rule Abduction.}
Attributes in the RAVEN dataset follow row-major binary or ternary relations \cite{zhang2019raven}. All possible binary $\mathbb{V} ^2$ and ternary $\mathbb{V} ^3$ relation pairs in the RPM test are presented in Table \ref{Table:RelationSets}, where $\boldsymbol{v}_{(i,j)}$ denotes the HD attribute representation for a given attribute in the context panel at row $i$ and column $j$, and $\boldsymbol{v}_{\left(y\right)}$ represents the corresponding HD attribute representation of the target answer panel. Consequently, rule abduction can be formulated as an optimization problem: For both numerical and logical rules, the rule abduction module must identify a set of operator powers $OP_{1:M}$ such that all $N$-ary ($N=2,3$) relation pairs yield the same output $\boldsymbol{r}_{Num/Lgc}$ when processed through their respective relation functions $R_{Num/Lgc}$. Formally:
\begin{equation}
\label{reasoningvaluefunction}
\underset{OP_{1:M}}{\max} s^N =\prod_{\mathbb{V} _{i}^{N},\mathbb{V} _{j}^{N}\in \mathbb{V} _{}^{N}}^{i\ne j}{sim\left( R\left( \mathbb{V} _{i}^{N},OP_{1:M} \right) ,R\left( \mathbb{V} _{j}^{N},OP_{1:M} \right) \right)}
\end{equation}
where $sim$ denotes cosine similarity, and $R$ represents either the numerical relation function $R_{Num}$ or the logical relation function $R_{Lgc}$.  $\mathbb{V} _{i}^{N},\mathbb{V} _{j}^{N}\in \mathbb{V} _{}^{N}$ ($i\ne j$) refers to any two relation pairs selected from $\mathbb{V}^N$ ($N=2,3$). $s^N_{Num/Lgc}$ represents the overall similarity between the outputs $r$ of all corresponding relation pairs and can be interpreted as an unnormalized probability of the corresponding rule.  

Based on the above idea, in the rule abduction module (Figure \ref{fig:ModelArchitecture}c Left), the HD attribute representations of $8$ context panels for a given numerical attribute are input into a trainable neural network $f_{\phi }^{Num}$ as the rule learner to predict the operator powers $\widehat{OP}_{1:M}$, which are expected to achieve the optimization of the objective defined in Equation \ref{reasoningvaluefunction}. Subsequently, all attribute sets for binary and ternary relations (Table \ref{Table:RelationSets}) are input into the corresponding numerical relation functions using the predicted $\widehat{OP}_{1:M}$, and their outputs $\hat{\boldsymbol{r}}_{Num}$ are obtained. Based on the outputs $\hat{\boldsymbol{r}}_{Num}$ from the relation functions, the unnormalized probability $s^2_{Num}$ and $s^3_{Num}$ for binary and ternary relations, respectively, can be computed (Equation \ref{reasoningvaluefunction}). The operator powers $\widehat{OP}_{1:M}$ and the averaged output $\bar{\boldsymbol{r}}_{Num}$ with larger $s^N_{Num}$ are then defined as the underlying numerical rule.

Similarly, another trainable rule learner, $f_{\varphi }^{Lgc}$, is used to predict the operator powers $\widehat{OP}_{1:M}$ for the logical rules associated with the attribute \textit{Position} (Figure \ref{fig:ModelArchitecture}c Right). Since the logical rules in the Raven dataset only involve ternary relations, all logical representations are organized according to the attribute sets of ternary relations outlined in Table \ref{Table:RelationSets}, and are then input into the logical ternary relation functions with the predicted parameters $\widehat{OP}_{1:M}$. Based on the outputs $\hat{\boldsymbol{r}}_{Lgc}$ from the relation functions, the unnormalized probability $s^3_{Lgc}$ for logical relations can be computed (Equation \ref{reasoningvaluefunction}). Subsequently, $s^3_{Lgc}$ is compared with $s^N_{Num}$, which corresponds to numerical relations for the attribute Position. If $s^3_{Lgc}$ is larger, the operator powers $\widehat{OP}_{1:M}$ and the averaged output $\bar{\boldsymbol{r}}_{Lgc}$ for logical relations can be interpreted as the underlying logical rule.

\textbf{Rule Execution.}
After obtaining the predicted operator powers ${\widehat{OP}_{1:M}}$ and the outputs $\bar{\boldsymbol{r}}_{Num/Lgc}$ that represent the rules, we apply these rules to infer the HD attribute representations of the missing panel. For a given attribute, the corresponding attribute representations from the first two panels in the third row of the RPM test are input into the inverse numerical and logical relation functions using the predicted ${\widehat{OP}_{1:M}}$ and $\bar{\boldsymbol{r}}_{Num/Lgc}$, resulting in the retrieval of the missing HD attribute representation $\hat{\boldsymbol{v}}_{\left( 3,3 \right)}$:
\begin{equation}
	\hat{\boldsymbol{v}}_{\left( 3,3 \right)}=R_{Num/Lgc}^{-1}\left( \mathbb{V} ,\widehat{OP}_{1:M},\bar{\boldsymbol{r}}_{Num/Lgc} \right) 
\end{equation}
where $\mathbb{V} =(\boldsymbol{v}_{\left( 3,2 \right)})$ for binary relation rules and $(\boldsymbol{v}_{\left( 3,1 \right)},\boldsymbol{v}_{\left( 3,2 \right)})$ for ternary relation rules. 

\textbf{The final answer selection.} 
Finally, we calculate the similarity between all HD attribute representations of the missing panel ($ \boldsymbol{v}^{position}_{\left( 3,3 \right)},\boldsymbol{v}^{number}_{\left( 3,3 \right)},\boldsymbol{v}^{type}_{\left( 3,3 \right)},\boldsymbol{v}^{size}_{\left( 3,3 \right)},\boldsymbol{v}^{color}_{\left( 3,3 \right)}  $) and the corresponding attribute representations of each candidate panel $\left( y \right)$. The predicted answer panel $\hat{y}$ is the one with the highest total similarity score.


\subsection{Model training}

\subsubsection{\textbf{End-to-end training}} During end-to-end training, the Rel-SAR model utilizes the rule labels provided by the RAVEN dataset and the answer labels to optimize the objectives of visual perception (Equation \ref{PerceptionValueFunction}) and rule abduction (Equation \ref{reasoningvaluefunction}). Based on the rule labels, the corresponding ground-truth $OP_{1:M}^{gt}$ and $\boldsymbol{r}_{Num/Lgc}^{gt}$, which represent the rules, can be obtained from Table \ref{Tab:RulesAndOP}. To facilitate the learning of $OP_{1:M}$, we design the loss function $\mathcal{L} _{op}$, which constrains the rule learners $f_{\phi }^{Num}$ and $f_{\varphi }^{Lgc}$ to optimize the objective described in Equation \ref{reasoningvaluefunction}:
\begin{equation}
	\mathcal{L}_{op}=MSE\left( \widehat{OP}_{1:M}, OP_{1:M}^{gt} \right)
\end{equation}
which is a mean square error ($MSE$) loss between the predicted operator powers $\widehat{OP}_{1:M}$ and the corresponding ground-truth. Additionally, we introduce the loss function $\mathcal{L} _{\boldsymbol{r}}$ to ensure consistent outputs when the inputs to the relation function follow a given rule. This is formulated as follows:
\begin{equation}
	\mathcal{L} _{\boldsymbol{r}}=\sum_i^{}{\left( 1-sim\left( \hat{\boldsymbol{r}}_i,\boldsymbol{r}^{gt} \right) \right)}
\end{equation}
where $\widehat{\boldsymbol{r}}_{i}$ denotes the output of the relation function for the $i$-th relation pair in $\mathbb{V} ^2$ and $\mathbb{V} ^3$. The overall loss function $\mathcal{L}$ for end-to-end training is constructed as: 
\begin{equation}
	\mathcal{L} =\sum{\mathcal{L} _{op}}+\sum{\mathcal{L} _{\boldsymbol{r}}}
\label{EQU:OverallLossForendtoend}
\end{equation}
where $\sum$ represents the sum of the loss functions across all attributes, for both binary and ternary relation functions, and for both numerical and logical rule types. By minimizing $\mathcal{L}_{op}$ and $\mathcal{L}_{\boldsymbol{r}}$ simultaneously during training, the optimization objective of the perception network $f_{\theta}$, as described in Equation \ref{PerceptionValueFunction}, can be achieved. This is because, under the constraints of numerous ground-truth rules ($OP_{1:M}^{gt}$ and $\boldsymbol{r}_{Num/Lgc}^{gt}$), the expected theoretical SHDR constructed from HD attribute representations in the codebooks will be a competitive representation, guiding the reasoning process toward optimality ($\mathcal{L}_{op} \rightarrow 0$ and $\mathcal{L}_{\boldsymbol{r}} \rightarrow 0$).

\subsubsection{\textbf{End-to-End Training with auxiliary attribute labels}}
Following previous work, we assess the performance of the Rel-SAR model in end-to-end training using both auxiliary attribute labels and answer labels. Here, a cosine similarity loss is employed as the perception loss function to enhance the similarity between the estimated SHDR $\hat{\mathcal{S}}^{ind}$ of the perception frontend and the theoretical SHDR $\mathcal{S}$ (Equation \ref{SHDR}) derived from attribute labels, thereby optimizing the trainable perception network $f_{\theta}$ (Equation \ref{PerceptionValueFunction}). The perception loss function $\mathcal{L}_p$ is defined as follows:
\begin{equation}
	\mathcal{L}_p =1-sim\left( \hat{\mathcal{S}}^{ind} ,\mathcal{S} \right) 
\label{Loss_Perception}
\end{equation}

Meanwhile, to achieve rule learning through the optimization objective described in Equation \ref{reasoningvaluefunction}, we introduce the loss function $\mathcal{L}_{rs}$ to increase the overall similarity between the outputs $\boldsymbol{r}$ of all corresponding relation pairs in $\mathbb{V} ^2$ and $\mathbb{V} ^3$ (Table \ref{Table:RelationSets}), respectively. This is formulated as follows:
\begin{equation}
	\mathcal{L} _{rs}=1-\prod_{i,j}^{i\ne j}{sim\left( \hat{\boldsymbol{r}}_i,\hat{\boldsymbol{r}}_j \right)}
	\label{Loss_Reasoning}
\end{equation}
where $\hat{\boldsymbol{r}}_i$ and $\hat{\boldsymbol{r}}_j$ denotes the outputs of relation functions for any two distinct sets of relation pairs in $\mathbb{V} ^2$ and $\mathbb{V} ^3$, including attribute sets from the answer panel $y$. Therefore, the overall loss function $\mathcal{L}$ for end-to-end training with auxiliary attribute labels is constructed as:
\begin{equation}
	\mathcal{L} =\sum{\mathcal{L} _{p}}+\sum{\mathcal{L} _{rs}}
	\label{EQU:OverallLossForendtoendattributelabels}
\end{equation}
where the former $\sum$ represents the sum of the loss functions across all context panels and the answer panel, while the latter $\sum$ represents the sum of the loss functions across all attributes.

Additionally, similar to other neuro-symbolic approaches, the perception frontend and reasoning backend in Rel-SAR can also be trained independently using the loss functions $\mathcal{L}_p$ and $\mathcal{L}_{rs}$, respectively.

\section{Experiments}

\subsection{Datasets \& Experiment setup}
We evaluate our model on the RAVEN \cite{zhang2019raven} and I-RAVEN \cite{hu2021stratified} datasets. Each dataset contains 70,000 RPM samples, which are divided into training, validation, and test sets with a 6:2:2 ratio. We use a ResNet-50 as the encoder ($f_{\theta}$) to map the image panels $\mathcal{X}$ to their SHDR $\mathcal{S} \in \mathbb{R} ^d$, where the dimension $d$ of all vectors is set to $3000$. Two 5-layer fully connected networks as rule learners ($f_{\phi }^{Num}$ and $f_{\varphi }^{Lgc}$) are employed to extract $OP_{1:M}$ from updated \textit{Position} representation and other attribute (\textit{Number}, \textit{Type}, \textit{Size} and \textit{Color}) representation, respectively. The existence vectors of the backend codebook are set to HRR because the normalization of FHRR in the superposition operation will invalidate the logical reasoning. All other vectors are FHRR. We utilize the AdamW optimizer with a learning rate of $1\times 10^{-4}$ and a weight decay of $1\times 10^{-4}$. 


\begin{table*}
	\centering
	\caption{Test accuracy (\%) on RAVEN and I-RAVEN Dataset}
	\scalebox{0.85}{
		\begin{tabular}{c|l|cccccccc|cccccccc}
			\toprule
			\multicolumn{2}{c}{\multirow{2}{*}{Method}} & \multicolumn{8}{c}{RAVEN} & \multicolumn{8}{c}{I-RAVEN} \\ \cmidrule(r){3-10} \cmidrule(r){11-18}
			\multicolumn{2}{c}{} & Avg & Center & 2X2 & 3X3 & L-R & U-D & O-IC & O-IG & Avg & Center & 2X2 & 3X3 & L-R & U-D & O-IC & O-IG \\ \midrule
			\multirow{11}{*}{\shortstack{Deep Learning\\Model}}
			& WReN\cite{barrett2018measuring} & 14.7 & 13.1 & 28.6 & 28.3 & 7.5  & 6.3 & 8.4 & 10.6  & 23.8 & 29.4 & 26.8 & 23.5 & 21.9 & 21.4 & 22.5 & 21.5 \\ 
			~ & LEN\cite{zheng2019abstract} & 72.9 & 80.2 & 57.5 & 62.1 & 73.5 & 81.2 & 84.4 & 71.5 & 41.4 & 56.4 & 31.7 & 29.7 & 44.2 & 44.2 & 52.1 & 31.7 \\ 
			~ & CoPINet\cite{zhang2019learning} & 91.4 & 95.1 & 77.5 & 78.9 & 99.1 & 99.7 & 98.5 & 91.4 & 46.1 & 54.4 & 36.8 & 31.9 & 51.9 & 52.5 & 52.2 & 42.8 \\ 
			~ & MXGNet\cite{wang2020abstract} & 84.0 & 94.3 & 60.5 & 64.9 & 96.6 & 96.4 & 94.1 & 81.3 & 33.1 & 40.7 & 27.9 & 24.7 & 35.8 & 34.5 & 36.4 & 31.6 \\ 
			~ & SCL\cite{wu2020scattering} & 91.6 & 98.1 & 91.0 & 82.5 & 96.8 & 96.5 & 96.0 & 80.1 & 95.0 & 99.0 & 96.2 & 89.5 & 97.9 & 97.1 & 97.6 & 87.7 \\ 
			~ & SRAN\cite{hu2021stratified} & 54.3 & --- & --- & --- & --- & --- & --- & ---  & 60.8 & 78.2 & 50.1 & 42.4 & 70.1 & 70.3 & 68.2 & 46.3 \\
			~ & Rel-Base\cite{spratley2020closer} & 91.7 & 97.6 & 85.9 & 86.9 & 93.5 & 96.5 & 97.6 & 83.8 & 91.1 & --- & --- & --- & --- & --- & --- & --- \\ 
			~ & MRNet\cite{benny2021scale} & 74.4 & 96.2 & 49.1 & 45.9 & 93.7 & 94.2 & 92.5 & 51.3 & 75.0 & 96.8 & 45.6 & 39.6 & 95.7 & 95.9 & 95.6 & 55.5 \\ 
			~ & DCNet\cite{zhuo2022effective} & 93.6 & 97.8 & 81.7 & 86.7 & 99.8 & 99.8 & 99.0 & 91.5 & 49.4 & 57.8 & 34.1 & 35.5 & 58.5 & 60.0 & 57.0 & 42.9 \\
			~ & STSN\cite{mondal2023learning} & 89.7 & --- & --- & --- & --- & --- & --- & ---  & 95.7 & 98.6 & 96.2 & 88.8 & 98.0 & 98.8 & 97.8 & 92.0 \\
			~ & PredRNet\cite{yang2023neural} & 95.8 & --- & --- & --- & --- & --- & --- & ---  & 96.5 & --- & --- & --- & --- & --- & --- & --- \\
			~ & DRNet\cite{zhao2024learning}  & 96.9 & --- & --- & --- & --- & --- & --- & ---& 97.6 & --- & --- & --- & --- & --- & --- & --- \\ \midrule
			\multirow{4}{*}{\shortstack{Neuro-Symbolic\\Model}} & PrAE \cite{zhang2021abstract} & 65.0 & 76.5 & 78.6 & 28.6 & 90.1 & 90.9 & 48.1 & 42.6 & 77.0 & 90.5 & 85.4 & 45.6 & 96.3 & 97.4 & 63.5 & 60.7 \\ 
			~ & ALANS\cite{zhang2022learningb} & 74.4 & 69.1 & 80.2 & 75.0 & 72.2 & 73.3 & 76.3 & 74.9 & 78.5 & 72.3 & 79.5 & 72.9 & 79.2 & 79.6 & 85.9 & 79.9 \\ 
			~ & NVSA\cite{hersche2023neuro} & 87.7 & 99.7 & 93.5 & 57.1 & 99.8 & 99.7 & 98.6 & 65.4 & 88.1 & 99.8 & 96.2 & 54.3 & 100 & 99.9 & 99.6 & 67.1 \\
			~ & Rel-SAR (Ours) & 96.5 & 99.1 & 95.7 & 96.2 & 99.6 & 99.6 & 99.1 & 86.2 & 98.0 & 99.8 & 97.1 & 98.1 & 100 & 100 & 99.9 & 90.9 \\ \midrule
			\multirow{4}{*}{\shortstack{Neuro-Symbolic\\Model\\(attribute labels)}} & NVSA\cite{hersche2023neuro} & 98.5 & 100 & 99.4 & 96.3 & 100 & 100 & 100 & 93.9 & 99.0 & 100 & 99.5 & 97.1 & 100 & 100 & 100 & 96.4 \\ 
			~ & Xu et al.\cite{xu2023abstract} & 92.9 & 98.8 & 91.9 & 93.1 & 99.2 & 99.1 & 98.2 & 70.1 & 93.2 & 99.5 & 89.6 & 89.7 & 99.7 & 99.5 & 99.6 & 74.7 \\
			~ & ALANS-V\cite{zhang2022learningb} & 94.4 & 98.4 & 91.5 & 87.0 & 97.3 & 96.4 & 97.3 & 93.2 & 93.5 & 98.9 & 85.0 & 83.2 & 90.9 & 98.1 & 99.1 & 89.5 \\ 
			~ & Rel-SAR (Ours) & 96.6 & 97.9 & 94.3 & 96.6 & 99.0 & 98.7 & 97.9 & 92.2 & 98.1 & 98.5 & 96.7 & 97.8 & 99.4 & 99.8 & 99.2 & 95.6\\ \bottomrule
	\end{tabular}}
	\label{main_result}
\end{table*}

\subsection{End-to-end learning results}

Here we first evaluate the Rel-SAR model on both RAVEN \cite{zhang2019raven} and I-RAVEN \cite{hu2021stratified} datasets using end-to-end learning. The results, summarized in Table \ref{main_result}, compare our model with both deep neural network methods and neuro-symbolic AI methods. Rel-SAR achieves an average accuracy of $96.5\%$ on RAVEN and $98.0\%$ on I-RAVEN, comparable to the previous best-performing deep network, DRNet ($96.9\%$ on RAVEN and $97.6\%$ on I-RAVEN).  Compared to the previous neuro-symbolic method NVSA, Rel-SAR demonstrates a significant accuracy improvement on configurations involving rules based on the \textit{position} attribute, including \textsf{2x2Grid} ($+2.2\%$), \textsf{3x3Grid} ($+39.1\%$), and \textsf{Out-InGrid} ($+20.8\%$). This results in an average accuracy improvement of $8.8\%$ on RAVEN and $9.9\%$ on I-RAVEN. These improvements are attributed to the effectiveness of our circular and logical HD representations for the \textit{position} attribute, as well as the numerical and logical relation functions used in rule abduction and execution. Additionally, Rel-SAR utilizes only the answer panel during training, without leveraging information from other candidate panels. This ensures that Rel-SAR does not exploit potential defects in the RAVEN dataset for shortcut learning \cite{hu2021stratified}, resulting in similar accuracy on both RAVEN and I-RAVEN datasets ($96.5\%$ vs. $98.0\%$).

In addition, since neuro-symbolic models are hybrid architectures that disentangle perception and reasoning, they are often trained with auxiliary attribute labels. We therefore evaluate the Rel-SAR model on both datasets when trained with additional attribute labels. The results, presented in the bottom section of Table \ref{main_result}, show that our model achieves an average accuracy of $96.6\%$ on RAVEN and $98.1\%$ on I-RAVEN. Among the neuro-symbolic models compared, ALANS Learner is the most similar to Rel-SAR, as both models incorporate learnable parameters in their reasoning backends. Compared to ALANS, our model demonstrates an average accuracy improvement of $2.2\%$ on RAVEN and $4.6\%$ on I-RAVEN. In contrast, the reasoning backends in NVSA and Xu's model rely on pre-designed rule templates or algebraic invariance modules without learnable parameters. NVSA, with its comprehensive rule template library, achieves near-perfect accuracy when trained with auxiliary attribute labels. Although our model slightly underperforms NVSA when trained with auxiliary attribute labels, it exhibits an important advantage: when trained without attribute labels, both NVSA and ALANS Learner suffer significant performance drops, whereas our model maintains nearly identical performance. This consistency highlights the robust synergy between the perception and reasoning modules in our model.

\subsection{Perception results learned with the attribute labels}

\begin{table}[!ht]
	\centering
	\caption{The visual perception accuracy on RAVEN dataset. }
	\scalebox{0.85}{
		\begin{tabular}{lllllllll}
			\toprule
			Method & Avg. & Center & 2X2 & 3X3 & L-R & U-D & O-IC & O-IG \\ \midrule
			PrAE & 85.27 & 88.65 & 93.56 & 73.95 & 100 & 100 & 94.23 & 46.25 \\ 
			Xu et al. & 96.10 & 100 & 100 & 99.99 & 100 & 100 & 99.96 & 72.78  \\ 
			NVSA & 99.76 & 100 & 99.83 & 98.61 & 99.97 & 99.96 & 99.97 & 99.95 \\ \midrule
			Rel-SAR & 99.99 & 100 & 100 & 99.99 & 100 & 100 & 100 & 99.94 \\ \bottomrule
		\end{tabular}
		\label{perception_acc}
	}
\end{table}

Similar to other neuro-symbolic methods, the perception frontend in Rel-SAR can also be independently trained with attribute labels using the perception loss function (Equation \ref{Loss_Perception}). By querying the estimated HD attribute representations for the highest similarity with the attribute vectors in the frontend codebooks, we can retrieve the predicted entity-level attributes. A panel’s features are considered correctly extracted only if the attributes of all objects in the panel match those provided by the dataset. As shown in Table \ref{perception_acc}, the evaluation results demonstrate that the perception frontend in our model achieves an average panel accuracy of $99.99\%$ on the RAVEN dataset, maintaining nearly perfect performance, consistent with NVSA. Notably, NVSA reports a resolution issue in the RAVEN dataset, where some objects in the inner regions of the \textsf{O-InGrid} configuration have a different size attribute but the same image representation\cite{hersche2023neuro}. To address this, we adopt a similar solution to NVSA, merging classes with different sizes but identical panel representation.

To evaluate the generalization capability of the perception frontend of our model, we adopt the experimental settings outlined in \cite{hersche2023neuro} to test the model on unseen combinations of attribute values. Specifically, we focus on the single-object case in the \textsf{2x2Grid} configuration. As shown in Table \ref{perception_gen_single}, we select two attributes (e.g. \textit{Position} and \textit{Type}) along with partial value sets for each (e.g. \textit{Position} $\in \{0,3\}$ and \textit{Type} $\in \{0,2\}$) as the target attributes. Panels containing these target attribute values are included in the training set, while panels lacking them are designated for the test set. This approach ensures that the attribute sets of the training and test datasets are entirely disjoint, enabling a rigorous evaluation of the module's ability to generalize beyond the observed data.

In this experiment, we observe that our perception frontend struggles to identify unseen combinations of attribute values when the value vectors $\boldsymbol{v}$ in Equation \ref{SHDR} are Random Vectors (RVs). We attribute this issue to the orthogonality of RVs, which hinders the model's ability to recognize adjacency concepts. To address this, we replace the RVs with fractional power encoding (FPE) vectors based on a Gaussian kernel, which introduces progressive similarity between vectors \cite{plate1994distributed, Frady_Kleyko_Kymn_Olshausen_Sommer_2021}. As shown in Table \ref{perception_gen_single}, our model demonstrates better generalization on unseen attribute combinations compared to NVSA \cite{hersche2023neuro} when using FPE vectors with progressive similarity (NVs). However, the type-size combination still shows low accuracy, likely because there is no clear continuous progression between adjacent attribute values of type (e.g., triangle and square).

\begin{figure}[!ht]
	\centering
	\includegraphics[scale=0.23]{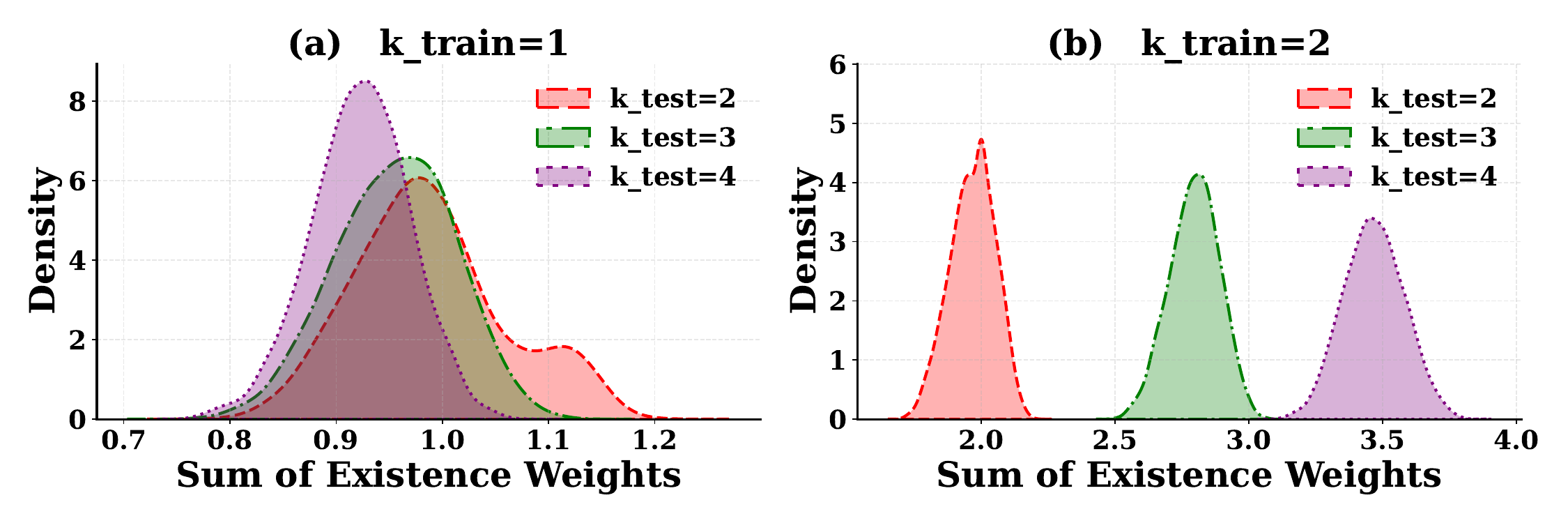}
	\caption{\textbf{Attention weights for \textit{existence} on \textsf{2X2Grid}.}
		For the \textit{existence} attribute, the presence weights ($W_{j}^{exist}\left( 1 \right)$ in Equation \ref{query_exist}) represent the probability of an object being present at the $j$th position. The sum of these weights provides an approximate indication of the total number of objects in a panel. \textbf{(a)} When $k_{train}=1$, the model consistently predicts the presence of only one object in the panel, regardless of the actual test scenario. \textbf{(b)} When $k_{train}=2$, the model learns to distinguish between panels with varying numbers of objects, effectively adapting its feature extraction process.}
	\label{Figure:weightanalyze}

\end{figure}

We also conduct an experiment to evaluate the generalization ability of our perception frontend when applied to an unseen number of objects\cite{hersche2023neuro}. The training set consists of panels with a finite number of objects (e.g., $k_{train}=1$), while the test set consists of panels with a larger number of objects (e.g., $k_{test}=2$). As shown in Table \ref{pecception_gen_more}, when $k_{train}=1$, our model fails to accurately predict the attributes of panels containing more objects. By analyzing the predicted attribute values, we observe that our model exhibits a consistent distribution of predicted attribute attention weights across different test sets, attributable to the limited diversity (Figure \ref{Figure:weightanalyze}a). However, as the number of objects in the training panels increases, the model learns more complex patterns and can correctly allocate attention to different positions (Figure \ref{Figure:weightanalyze}b). Our model achieves perfect generalization performance when $k_{train}=2$ for \textsf{2x2Grid} and when $k_{train}=3$ for \textsf{3x3Grid}.

\begin{table*}[!ht]
	\centering
	\caption{Accuracy of attribute-value generalization on \textsf{2x2Grid} containing $k=1$ object}
	\begin{tabular}{ccccccc}
		\toprule
		Training combinations  & Testing combinations & Train sample & Test Sample & NVSA & Rel-SAR (RV) & Rel-SAR (NV) \\ 
		\midrule
		Position $\in \{0,3\}$ OR & Position $\notin \{0,3\}$ AND & \multirow{2}{*}{6720} &  \multirow{2}{*}{2880} &  \multirow{2}{*}{0.0} &  \multirow{2}{*}{0.0}  & \multirow{2}{*}{26.8} \\
		Type $\in \{0,2\}$  & Type $\notin \{0,2\}$  \\ \midrule
		Position $\in \{0,3\}$ OR & Position $\notin \{0,3\}$ AND & \multirow{2}{*}{6400} &  \multirow{2}{*}{3200} & \multirow{2}{*}{15.1}  &  \multirow{2}{*}{0.0}&\multirow{2}{*}{31.4} \\
		Size $\in \{1,5\}$  & Size $\notin \{1,5\}$  \\ \midrule
		Position $\in \{0,3\}$ OR & Position $\notin \{0,3\}$ AND & \multirow{2}{*}{6720} &  \multirow{2}{*}{2880} & \multirow{2}{*}{34.8} &  \multirow{2}{*}{0.0} & \multirow{2}{*}{73.5} \\
		Color $\in \{0,3,6,8\}$  & Color $\notin \{0,3,6,8\}$  \\ \midrule
		Type $\in \{0,2\}$ OR & Type $\notin \{0,2\}$ AND & \multirow{2}{*}{5760} &  \multirow{2}{*}{3840} & \multirow{2}{*}{0.0} &  \multirow{2}{*}{0.0}  & \multirow{2}{*}{13.9} \\
		Size $\in \{1,5\}$  & Size $\notin \{1,5\}$  \\ \midrule
		Type $\in \{0,2\}$ OR & Type $\notin \{0,2\}$ AND & \multirow{2}{*}{6144} &  \multirow{2}{*}{3456} & \multirow{2}{*}{72.0} &  \multirow{2}{*}{0.0} &  \multirow{2}{*}{89.5} \\
		Color $\in \{0,3,6,8\}$  & Color $\notin \{0,3,6,8\}$ \\ \midrule
		Size $\in \{1,5\}$ OR & Size $\notin \{1,5\}$ AND & \multirow{2}{*}{5760} &  \multirow{2}{*}{3840} & \multirow{2}{*}{29.3} & \multirow{2}{*}{0.0}  & \multirow{2}{*}{82.6} \\
		Color $\in \{0,3,6,8\}$  & Color $\notin \{0,3,6,8\}$ \\ 
		\bottomrule
	\end{tabular}
	\label{perception_gen_single}
\end{table*}

\begin{table*}[ht]
	\centering
	\caption{Generalization to a growing number of unseen objects in the RAVEN panel}
	\begin{tabular}{lllllllllllll}
		\toprule
		Training  & Training& \multirow{2}{*}{$k_{train}$} & \multicolumn{8}{c}{$k_{test}$} & \multirow{2}{*}{Avg} & ODD \\ \cmidrule{4-11}
		Configuration & Samples & ~ & 2 & 3 & 4 & 5 & 6 & 7 & 8 & 9 &  & Avg \\ \midrule
		\textsf{2x2Grid} & 9600 & 1 & 0.3 & 0.0 & 0.0 & --- & --- & --- & --- & --- & 0.1 & 0.1 \\ 
		\textsf{2x2Grid} & 19200 & 2 & 100.0 & 100.0 & 100.0 & --- & --- & --- & --- & --- & 100.0 & 100.0 \\\midrule
		\textsf{3x3Grid} & 21600 & 1 & 0.2 & 0.0 & 0.0 & 0.0 & 0.0 & 0.0 & 0.0 & 0.0 & 0.0 & 0.0 \\ 
		\textsf{3x3Grid} & 43200 & 2 & 100.0 & 100.0 & 99.4 & 70.3 & 21.6 & 0.4 & 0.0 & 0.0 & 49.0 & 41.7 \\
		\textsf{3x3Grid} & 64800 & 3 & 100.0 & 100.0 & 100.0 & 100.0 & 100.0 & 100.0 & 100.0 & 100.0 & 100.0 & 100.0 \\
		\textsf{3x3Grid} & 86400 & 4 & 100.0 & 100.0 & 100.0 & 100.0 & 100.0 & 100.0 & 100.0 & 100.0 & 100.0 & 100.0 \\\bottomrule
	\end{tabular}
	\label{pecception_gen_more}
\end{table*}

%
%

\subsection{Reasoning results utilizing the attribute labels}

We analyze the performance of the reasoning backend of our model on the I-RAVEN dataset. Following the assumption of perfect perception in prior works \cite{camposampiero2024towards, hersche2024probabilistic}, we also utilize the ground truth attribute labels provided by the I-RAVEN dataset to generate HD attribute representations. The model is trained using the reasoning loss function (Equation \ref{Loss_Reasoning}), and the corresponding evaluation results are shown in Table \ref{reasoning_result}. We compare our model with LLM based GPT-3\cite{brown2020language}, deep neural network SCL\cite{wu2020scattering}, neuro-symbolic based method PrAE\cite{zhang2021abstract} and NVSA\cite{hersche2023neuro}, VSA based method Learn-VRF\cite{hersche2024probabilistic} and ARLC\cite{camposampiero2024towards}. The performance of GPT-3 is reported in \cite{hersche2024probabilistic}.  Our method achieves an accuracy of $99.2\%$, representing a $6.8\%$ improvement over the previous state-of-the-art model, ARLC. Notably, our approach significantly outperforms prior VSA-based methods ARLC on the \textsf{2X2Grid} ($+14.9\%$), \textsf{3X3Grid} ($+17.7\%$), and \textsf{O-InGrid} ($+9.4\%$) configurations, which involve rules on the \textit{Position} attribute. This improvement demonstrates that the circular and logical HD representations of \textit{Position}, along with the corresponding relation functions, effectively handle the reasoning of \textit{Arithmetic} and \textit{Progression} rules of \textit{Position}.

\begin{table*}
	\centering
	\caption{In-distribution accuracy on the I-RAVEN dataset}
	\begin{tabular}{lllllllll}
		\toprule
		Method & Avg & Center & 2X2 & 3X3 & L-R & U-D & O-IC & O-IG \\ \midrule
		GPT-3\cite{brown2020language} & 86.5 & 86.4 & 83.2 & 81.8 & 83.4 & 84.6 & 92.8 & 93.0 \\ 
		SCL\cite{wu2020scattering}  & 84.3 & 99.9 & 68.9 & 43.0 & 98.5 & 99.1 & 97.7 & 82.6 \\ 
		PrAE\cite{zhang2021abstract} & 71.1 & 83.8 & 82.9 & 47.4 & 94.8 & 94.8 & 56.6 & 37.4 \\ 
		NVSA\cite{hersche2023neuro}& 88.1 & 99.8 & 96.2 & 54.3 & 100 & 99.9 & 99.6 & 67.1 \\ 
		Learn-VRF\cite{hersche2024probabilistic} & 79.5 & 97.7 & 56.3 & 49.9 & 94.0 & 95.6 & 98.3 & 64.8 \\ 
		ARLC\cite{camposampiero2024towards} & 92.4 & 98.4 & 83.4 & 80.0 & 98.7 & 98.4 & 98.8 & 89.4 \\ \midrule
		Rel-SAR & 99.2 & 99.9 & 98.3 & 97.7 & 100.0 & 99.9 & 100.0 & 98.8 \\ 
		\bottomrule
	\end{tabular}
	\label{reasoning_result}
\end{table*}
\begin{table*}[ht]
	\centering
	\caption{Out-of-distribution accuracy on unseen rule-attribute pairs on I-RAVEN}
	\begin{tabular}{l lll llll llll}
		\toprule
		~ & \multicolumn{3}{c}{Type} & \multicolumn{4}{c}{Size} &  \multicolumn{4}{c}{Color} \\ \cmidrule(r){2-4} \cmidrule(r){5-8} \cmidrule(r){9-12}
		~ & Const. & Progr. & Dist.3 & Const. & Progr. & Arth. & Dist.3 & Const. & Progr. & Arth.  & Dist.3 \\ \midrule
		GPT-3\cite{brown2020language} & 88.5 & 86.0 & 88.6 & 93.6 & 93.2 & 92.6 & 71.6 & 94.2 & 94.7 & 94.3 & 65.8 \\ 
		Learn-VRF\cite{hersche2024probabilistic} & 100 & 100 & 99.7 & 100 & 100 & 99.8 & 99.8 & 100 & 98.8 & 100 & 100 \\ 
		ARLC\cite{camposampiero2024towards} & 100 & 98.6 & 99.7 & 100 & 100 & 99.6 & 99.6 & 100 & 100 & 100 & 99.8 \\ \midrule
		Rel-SAR & 99.7 & 99.6 & 100 & 99.3 & 99.6 & 97.9 & 100 & 96.8 & 89.5 & 86.4 & 70.7\\ \bottomrule
		
	\end{tabular}
	\label{OOD_reason}
\end{table*}

We also evaluate the out-of-distribution (OOD) generalization ability of our reasoning backend, following the experimental setup described in \cite{hersche2024probabilistic}. In this evaluation, a specific rule (e.g., \textit{Type}) for an attribute (e.g., \textit{Const.}) is designated as the target rule. The model is trained and validated using samples that exclude the target rule and is then tested on samples containing only the target rule. This experiments is conducted on the \textsf{Center} configuration, while training, validation, and test sets are filtered from the I-RAVEN dataset. The corresponding evaluation results are shown in Table \ref{OOD_reason}. For attributes \textit{Type} and \textit{Size}, our model, like Learn-VRF \cite{hersche2024probabilistic} and ARLC \cite{camposampiero2024towards}, demonstrates a near-perfect ability to generalize to unseen attribute rules. This capability arises from the use of unified HD vectors to represent values across different attributes, facilitating rule transfer between attributes. However, our model exhibits relatively limited performance on the \textit{Color} attribute. This could be due to the fact that \textit{Color} has a broader range of attribute values, making it difficult to transfer rules learned from attributes with narrower value ranges, such as \textit{Type} (0-4) and \textit{Size} (0-5), to the \textit{Color} (0-9) attribute. 


\section{Conclusion and Future Directions}

In this work, we propose Rel-SAR, a novel model that leverages VSA algebra to facilitate systematic rule abduction and execution. Rel-SAR adopts a neuro-symbolic architecture, where the perception frontend extracts diverse high-dimensional attribute representations with intrinsic algebraic properties, and the reasoning backend systematically derives a variety of rules based on relation functions. Extensive experiments demonstrate that Rel-SAR achieves superior performance while offers better interpretability and transparency on RPM tasks.

The perception frontend of Rel-SAR effectively extracts object-level attributes while preserving the structure information of the image through the binding and bundling operations of VSA.  By utilizing fractional power encoding vectors with progressive similarity, our model demonstrates its ability to generalize beyond the
observed data. However, Rel-SAR exhibits relatively low accuracy on the \textsf{Out-InGrid} configuration, which may be attributed to the small object size and resolution issue. As noted in \cite{yang2023neural}, CNNs with large kernel sizes or more stacked blocks are less effective at extracting features from RPM images with smaller objects. NVSA further shows that reducing the stride from 2 to 1 in the first convolutional block and removing the max-pooling layer in ResNet-18 can improve accuracy \cite{hersche2023neuro}. We believe that using CNNs with smaller kernel sizes and strides in the shallow layers will enhance the accuracy on the O-IG configuration.

Consistent with other neuro-symbolic methods, we also introduce auxiliary rule labels during training. This is because, without precise rule-driven guidance, the model struggles to learn meaningful structured high-dimensional representations (SHDR) of attributes. We note that slot attention \cite{locatello2020object} enables unsupervised scene decomposition, while VQ-VAE\cite{van2017neural} learns discrete latent representations, disentangling different concepts (i.e., attributes in RPM) from raw images. Therefore, we suggest employing a learnable frontend codebook, combined with slot attention to enable unsupervised extraction of SHDR from raw images in the future work.

The reasoning backend of Rel-SAR implements sytematic abductive reasoning based on diverse HD attribute representation and relation function. Our model has a significant improvement on configurations with rules on \textit{Position}. This demonstrates the effectiveness of circular and logical representations of \textit{Position} attribute. The reasoning backend of the Rel-SAR exhibits limited out-of-distribution (OOD) generalization ability. This limitation may stem from the rule learner, a multi-layer fully connected neural network, which lacks the capacity to generalize to OOD attribute values. A growing body of recent work emphasizes reasoning based on relations between perceptual inputs rather than the features of individual inputs\cite{webb2020emergent,altabaa2023abstractors, kerg2022neural}. This trend is encapsulated by the "relational bottleneck" principle\cite{webb2024relational}, which aims to mitigate the influence of the absolute size of attribute values on relational reasoning. However, relational bottleneck may struggle with more complex relations. Future studies can focus on combining relational bottlenecks with VSA algebra to handle complex relations while maintaining robust out-of-distribution generalization.


\bibliographystyle{IEEEtran}
\bibliography{Bib}

\appendices
\section{The full version of logical relation function}

The RPM-style logic rules are expressed as that the set of attribute values in the third panel of a row or column corresponds to the logic operation applied to the first two panels. The simplified version ($M=3$) of the logical relation function, $R_{Lgc}$, cannot handle all RPM-style logic rules, such as XOR. To address this limitation, we developed a full version of $R_{Lgc}$ with $OP_{1:M}$ ($M=5$) to describe all RPM-style logical rules that are meaningful, defined as follows:
\begin{equation}
	\begin{aligned}
		\boldsymbol{r}_{Lgc}&=R_{Lgc}\left( \boldsymbol{v}_{1:N},OP_{1:M} \right)\\
		 &=\left(\left( op_1\boldsymbol{v}_1\land op_2\boldsymbol{v}_2 \right) \lor \left( op_3\boldsymbol{v}_1\land op_4\boldsymbol{v}_2 \right)\right) \circ op_5\boldsymbol{v}_3
	 \end{aligned}
\end{equation}
where $op_i\in \{0,1\}$ determines whether to negate $\boldsymbol{v}_i$, with negation ($\lnot$) applied when $op_i=1$ and no negation applied when $op_i=0$. The implementation of $op_i\boldsymbol{v}_i$ is as follows:
\begin{equation}
    op_i\boldsymbol{v}_i = \left(\boldsymbol{e}\left(1\right)\right)^{\left(\circ op_i\right)}\circ\boldsymbol{v}_i
\end{equation}
where $\boldsymbol{e}\left(1\right)$ is BVs representing a Boolean value indicating \textit{True}. When $op_i=1$, $\left(\boldsymbol{e}\left(1\right)\right)^{\left(\circ op_i\right)}$ simplifies to $\boldsymbol{e}\left(1\right)$, which signifies the negation of HD vector $\boldsymbol{v}_i$ (see Table \ref{logicoperation}). Similarly, When $op_i=0$, $\left(\boldsymbol{e}\left(1\right)\right)^{\left(\circ op_i\right)}$ simplifies to $\boldsymbol{e}\left(0\right)$, which does not change the $\boldsymbol{v}_i$.

The inverse logical relation function is defined as follows:
\begin{table}[ht]
	\centering
	\caption{Logical rules and Corresponding combinations of $OP_{1:M}$ and $r$ in Logical relation functions}
	\label{logic_m5}
	\begin{tabular}{lcccccc}
		\toprule
		& $op_1$ & $op_2$ & $op_3$ & $op_4$ & $op_5$ & $\boldsymbol{r}$ \\ \midrule
		AND & 0 & 0 & 0 & 0 & 0 & 0 \\
		OR & 1 & 1 & 1 & 1 & 1 & 0\\ 
		DIFF & 0 & 1 & 0 & 1 & 0 & 0\\
		XOR & 0 & 1 & 1 & 0 & 0 & 0\\  
		\bottomrule
	\end{tabular}
\end{table}

\begin{equation}
	\begin{aligned}
		\boldsymbol{v}_N&=R_{Lgc}^{-1}\left( \boldsymbol{v}_{1:N-1},OP_{1:M},\boldsymbol{r} \right)\\
		&=op_5\left(\left( op_1\boldsymbol{v}_1\land op_2\boldsymbol{v}_2 \right) \lor \left( op_3\boldsymbol{v}_1\land op_4\boldsymbol{v}_2 \right)\right)\\
	\end{aligned}
\end{equation}

Several typical logical rules, along with the corresponding $OP_{1:M}$ and $\boldsymbol{r}$ in the full version of logical relation function, are presented in Table \ref{logic_m5}. The DIFF logic rule, also known as the \textit{Arithmetic}- rule for the \textit{Position} attribute of the RAVEN dataset, specifies that the set of attribute values for the third panel is the difference set of the attribute values from the first two panels. OR rule is the \textit{Arithmetic}+ rule for \textit{Position} attributes in RAVEN. For the OR, AND, and DIFF rules, $op_3$ and $op_4$ are equal of $op_1$ and $op_2$, respectively, allowing them to be represented using the simplified version $R_{Lgc}$. However, $OP_{1:2}$ and $OP_{3:4}$ of the XOR rule are different, making it impossible to express using the simplified version $R_{Lgc}$. Because the RAVEN dataset only involve \textit{Arithmetic}+ and \textit{Arithmetic}- logical rules, this work employs the simplified logical relation function.

\label{sec:appendix_1}

\end{document}